\documentclass[10pt,journal,compsoc]{IEEEtran}

\ifCLASSOPTIONcompsoc
  \usepackage[nocompress]{cite}
\else
  \usepackage{cite}
\fi
\ifCLASSINFOpdf
  \usepackage[pdftex]{graphicx}
\else
\fi
\usepackage{amsmath}
\usepackage{mathtools}
\usepackage{amssymb}
\usepackage{amsfonts}
\usepackage{soul}
\usepackage{color}
\usepackage{graphicx}
\usepackage{subcaption}
\usepackage{lipsum}
\usepackage{cases}

\usepackage{algorithm}
\usepackage{algorithmic}
\begin{document}
%
\newcommand{\sys}{\texttt{Fumos}\xspace}
\title{FedMS: Federated Learning with Mixture of Sparsely Activated Foundations Models}
%
%
%

\author{Panlong~Wu,~\IEEEmembership{}
        Kangshuo~Li,~\IEEEmembership{}
        Ting~Wang,~\IEEEmembership{}
        and~Fangxin~Wang,~\IEEEmembership{Member,~IEEE}
\IEEEcompsocitemizethanks{\IEEEcompsocthanksitem Panlong Wu is with the Future Network of Intelligence Institute and the School of Science and Engineering, The Chinese
University of Hong Kong, Shenzhen, Shenzhen 518172, China. E-mail: panlongwu@link.cuhk.edu.cn.
\IEEEcompsocthanksitem Kangshuo Li is with the School of Data Science, The Chinese
University of Hong Kong, Shenzhen, Shenzhen 518172, China. E-mail: 24ganbatte@gmail.com.
\IEEEcompsocthanksitem Ting Wang is with the School of Science and Engineering, The Chinese
University of Hong Kong, Shenzhen, Shenzhen 518172, China. E-mail: 011tingwang@gmail.com.
\IEEEcompsocthanksitem Fangxin Wang is with the School of Science and Engineering and the
Future Network of Intelligence Institute, The Chinese University of Hong
Kong, Shenzhen and Guangdong Provincial Key Laboratory of Future
Networks of Intelligence. Email: wangfangxin@cuhk.edu.cn.}
\thanks{Manuscript received xxx; revised xxx.}}

%
%

\markboth{IEEE TRANSACTIONS ON MOBILE COMPUTING}
{Shell \MakeLowercase{\textit{et al.}}: Bare Advanced Demo of IEEEtran.cls for IEEE Computer Society Journals}
%



\IEEEtitleabstractindextext{%
\begin{abstract}
Foundation models have shown great success in natural language processing, computer vision, and multimodal tasks. FMs have a large number of model parameters, thus requiring a substantial amount of data to help optimize the model during the training. Federated learning has revolutionized machine learning by enabling collaborative learning from decentralized data while still preserving the data privacy of clients. Despite the great benefits foundation models can have empowered by federated learning, they face severe computation, communication, and statistical challenges. In this paper, we propose a novel two-stage federated learning algorithm called FedMS. A global expert is trained in the first stage and a local expert is trained in the second stage to provide better personalization.  We construct a Mixture of Foundation Models (\texttt{MoFM}) with these two experts and design a gate neural network with an inserted gate adapter that joins the aggregation every communication round in the second stage. To further adapt to edge computing scenarios with limited computational resources, we design a novel Sparsely Activated LoRA (\texttt{SAL}) algorithm that freezes the pre-trained foundation model parameters inserts low-rank adaptation matrices into transformer blocks and activates them progressively during the training. We employ extensive experiments to verify the effectiveness of FedMS, results show that FedMS outperforms other SOTA baselines by up to  $55.25\%$ in default settings.

\end{abstract}

\begin{IEEEkeywords}
Federated Learning, Foundation Model, Edge Computing
\end{IEEEkeywords}}

\maketitle

\IEEEdisplaynontitleabstractindextext

%
\IEEEpeerreviewmaketitle

\ifCLASSOPTIONcompsoc
\IEEEraisesectionheading{\section{Introduction}\label{sec:introduction}}
\else
\section{Introduction}
\label{sec:introduction}
\fi
Foundation Model (FM) has emerged as a potent solution to address the growing demand for machine learning services. It presents several advantages over its predecessors, the traditional smaller models. FM stands out primarily due to its extensive number of parameters, surpassing the capacity of earlier models. This massive increased parameter space allows FM to capture intricate patterns and relationships in the data, resulting in improved performance across various machine learning tasks.

FM follows a distinct training methodology compared to smaller models. While smaller models often rely on task-specific training, FM employs a pre-training and fine-tuning strategy. Pre-training with large datasets allows FM to acquire a broad data understanding and representation learning capabilities. This pre-training phase acts as a stepping stone, equipping FM with substantial knowledge and context from diverse data sources. 
Consequently, when fine-tuning FM for specific tasks, they derive significant advantages from the initial pre-training, leading to enhanced performance across a diverse set of tasks.

Federated learning (FL) has revolutionized the landscape of machine learning by enabling the collaborative training of a shared model across multiple edge devices without the need to share raw data. By adopting FL, we can utilize the distributed edge data while preserving data privacy and overcoming the limitations of centralized training approaches. This collaboration allows for collective learning from diverse datasets while respecting user privacy and local data ownership.

FMs with a tremendous number of parameters are data-hungry for the reason that they have a large parameter space to be optimized during the training. By combining the power of FM with the decentralized approach of FL, we can leverage FL by allowing decentralized data from different sources to be used, thus enabling the enhanced generalization ability of FM. Each device in the federation contributes its local knowledge and data patterns to the training process, resulting in a more comprehensive and diverse understanding of the data. 
This combination allows us to harness the benefits of both FL's collaborative training across edge devices while preserving data privacy and FM's large parameter capacity and pre-training strategy. 
\textbf{Several challenges arise in the domain of FL with FM which make FL with FM hard to employ in real-world applications.} 

\begin{itemize}
    \item The first challenge lies within the substantial number of parameters possessed by FMs, which distinguishes them from traditional FL models that possess much fewer parameters. This differentiation introduces impediments in the areas of communication and networking, as the transmission of parameters of these FMs is significantly time-consuming for modern mobile networks.
    \item The second challenge arises from the huge computational resource requirements posed by FMs, particularly for edge devices with limited computing resources. The considerable computational costs associated with these FMs create difficulties in implementing FL on resource-constrained edge devices.
    \item The third challenge is that FL encounters statistical challenges due to non-IID decentralized data, potentially resulting in issues such as parameter divergence and data distribution biases which can significantly harm the performance. 
\end{itemize}

To fill the gap, this paper addresses the challenges associated with FL with FM by introducing FedMS algorithm, an FL algorithm with a Mixture of Foundations Models that have Sparsely Activated Parameters. The proposed FedMS algorithm consists of two training stages.

In the first training stage, each client owns a foundation model, and low-rank adaption matrices are inserted into every transformer block of the foundation model. During the training, the pre-trained weights of the foundation are frozen, and all the parameters of inserted matrices are activated to better extract global information. In each communication round only the inserted matrices join the weight aggregation to reduce bandwidths consumption. The trained foundation model in the first stage will be frozen in the second stage and act as the global expert.

In the second training stage, we for the first time form a Mixture of Foundation Models (\texttt{MoFM}) system in FL which specifically addresses the statistical challenges encountered in FL. We leverage the foundation model trained in the first stage as a global expert and introduce another local expert which is a foundation model initialized from the weights of the global model to provide better personalization. We design a gate model with a specially designed gate adapter inserted into it so that it can quickly adapt to the change of relationship between two experts and intelligently assign weights to the final decision of two experts. In each communication round, only the gate adapter's activated parameters join the aggregation to save communication resources.
To further tackle the computation challenges, we propose a Sparsely Activated LoRA (\texttt{SAL}) algorithm to activate inserted low-rank adaptation matrices in a progressive way through a controller to suit different edge resource conditions.




\textbf{In summary, the main contributions of this paper can be summarized as follows:}
\begin{itemize}
    \item We propose a communication and computation friendly two-stage personalized FL algorithm FedMS, which can capture the global feature information through collaborative learning and capture the local feature information through personalized learning.
    \item We propose a Sparsely Activated LoRA (\texttt{SAL}) algorithm that sparsely activates the trainable low-rank decomposition matrices injected into foundation models in a progressive way through a self-defined controller 
    to adapt to scarce computation and communication resources in edge computing scenarios.
    \item We propose a Mixture of Foundation Models (\texttt{MoFM}) algorithm which is, to our best knowledge the first work to construct a mixture of vision language foundation models in personalized federated learning to tackle the data heterogeneous in federated learning and we further prove the effectiveness of FedMS through extensive experiments.
    
\end{itemize}

\section{Background and Related Work}
\subsection{Foundation Model}
Recently, FMs have achieved remarkable success in various domains such as natural language processing, computer vision, and multimodal tasks. By utilizing deep learning techniques like self-supervised Learning and contrastive learning, FMs with a massive number of model parameters are trained on large datasets. Consequently, these models exhibit strong generalization, feature extraction, and comprehension abilities.

\textbf{Various works have been done related to FMs in natural language processing.}
Bert\cite{kenton2019bert}, referred to as Bidirectional Encoder Representation from Transformers, is an advanced natural language processing model introduced by Devlin et al. (2018). This model employs a transformer architecture and is pre-trained on extensive text data, using a masked language model pre-training objective. GPT-3\cite{brown2020language} is trained using a language modeling pre-training objective. By making the model do the next token prediction, it can utilize the massive unlabeled data from the internet and have a powerful few-shots learning ability.

\textbf{There are also various works that have been done related to visual language FMs.}
Contrastive Language Image Pre-training (CLIP)\cite{radford2021learning}, as a famous FM proposed by OpenAI. This model a visual encoder and a text encoder to extract the semantic meaning and encode images and texts into image features and text features. Throughout the training process, contrastive learning is employed to maximize the similarity between related images and texts while minimizing the similarity between unrelated ones.
DALL-E 3\cite{betker2023improving} is a modern text-to-image system that has extraordinary prompt-following capability. It addresses the noisy and inaccurate image captions issue by training another specially designed image captioner. 

\begin{figure*}[t]
   \centering
   \includegraphics[width=1\textwidth]{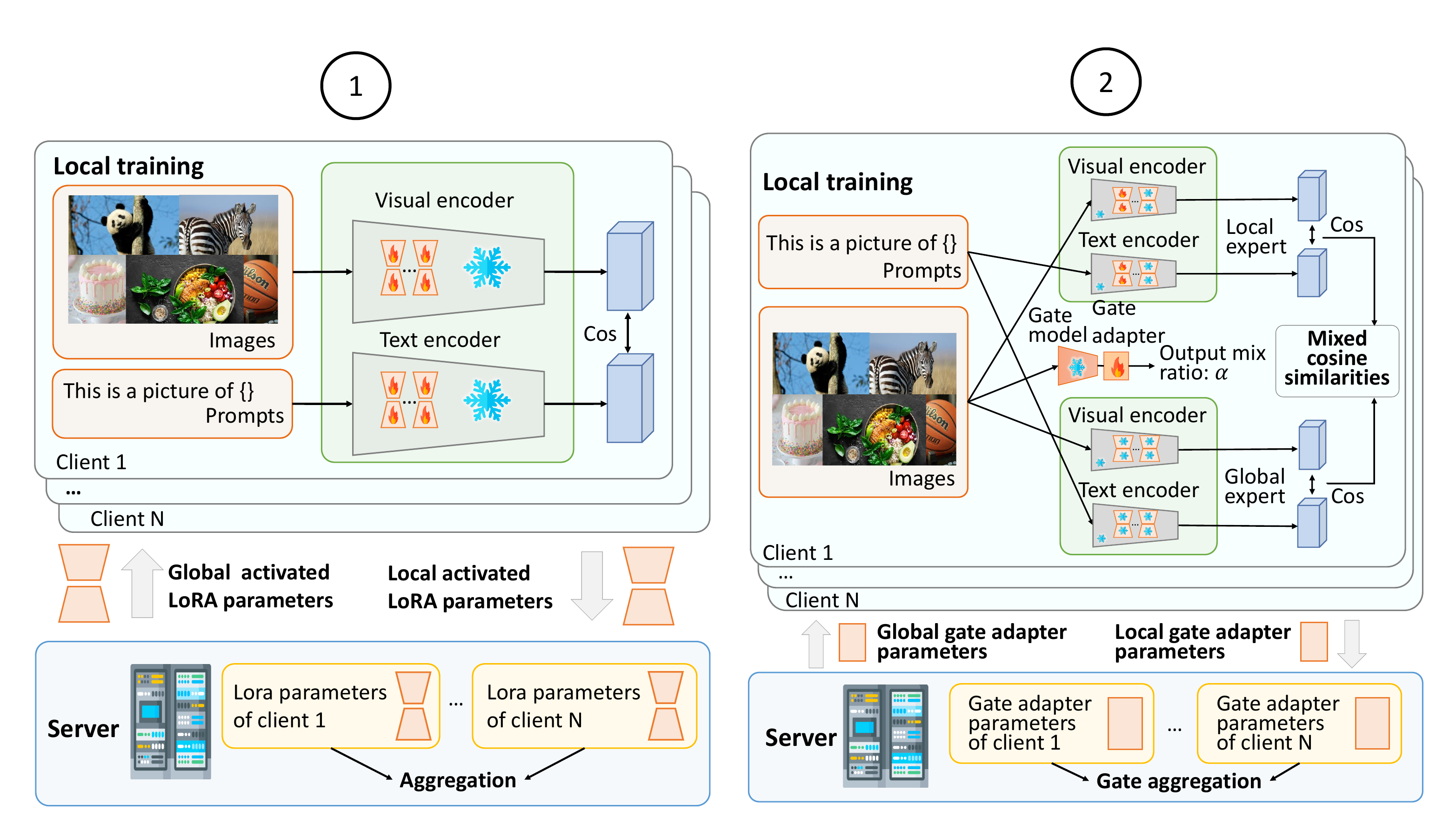}
   \caption{{FedMS Workflow}}\label{fig:FedMS}
\end{figure*}

\subsection{Federated Learning}
Federated learning \cite{mcmahan2017communication} is a machine learning technique that enables the training of decentralized data while preserving the privacy of clients participating in the training. Typically, every client doesn't share their data but their private model after local training in each communication round. Despite that FL has shown great potential in the Internet of Things, financial field, and smart healthcare, it still faces lots of challenges.

\textbf{Many studies focus on solving the statistical challenges.}
FL faces serious statistical challenges because the data distribution of the datasets is often non-iid which can lead to weight divergence after model aggregation.
Li et al. \cite{li2020federated} propose FedProx that handles the system heterogeneity by introducing an additional proximal term to prevent the local model updates from being far from the global model and thus can safely aggregate the local updates in statistical heterogeneity conditions. 
Li et al. \cite{li2021model} introduce MOON which uses the idea of contrastive learning to compare the representation learned by local models and the global model. Inspired by the philosophy of the global model has better feature extraction ability than local models that are trained on skewed local datasets. 
Zhang et al. \cite{zhang2022federated} design FedLC that introduces a fine-grained calibrated cross-entropy loss to mitigate the local gradient deviation and gives theoretical proof of the deviation bound after calibration. 
Zec et al. \cite{zec2020specialized} propose a personalized federated learning algorithm with a mixture of experts with a training pipeline of global expert training, local expert training, and mixer training.

\textbf{The communication efficiency issue is also an important issue that many researchers focus on.} MAO et al. \cite{mao2022communication} propose an Adaptive Quantized Gradient (AQG) algorithm to decide the level of quantization according to the gradient update of heterogeneous clients.
Huang et al. \cite{huang2020rpn} propose a Residual Pooling Network 
(RPN) based on the approximation of parameters and selection of parameters, and apply it to a CNN-based model FL training. Haddadpour et al. \cite{haddadpour2021federated} introduce an algorithm with periodical compressed communication. Specifically, they introduce the FedCOM algorithm to tackle the homogeneous client situation and the FedCOMGATE algorithm to tackle heterogeneous client situations. Chen et al. \cite{9916128} propose a federated learning algorithm that considers the weight quantization in wireless transmission and formulate the federated learning problem into a mixed-integer programming problem. Zhang et al. \cite{9928395} introduce a CFEL algorithm that jointly considers cloud-based federated learning and edge-based federated learning. Qu et al. \cite{9439928} design a partially Synchronized federated learning algorithm to accelerate the federated learning training.

\subsection{FL with FM}
Not many works have been done related to FL with FM. Zhang et al. \cite{zhang2023federated} propose a federated generative learning framework to utilize FM on the server to generate synthesized images given the transmitted prompt from the clients to improve the training performance of the model. Tao et al. \cite{guo2023promptfl} propose a PromptFL algorithm to replace the model aggregation in traditional FL to prompt aggregation to reduce communication and computation costs. Cai et al. \cite{cai2023efficient} design a AdaFL algorithm to fine-tune FMs for modern natural language processing tasks by inserting adapters into models, dividing clients into three groups, and observing each group's training accuracy to decide the best configuration for adapters. Lu et al. \cite{lu2023fedclip} propose a FedCLIP algorithm to insert adapters in the visual encoder of FM CLIP and test it on datasets in different domains. Zhang et al. \cite{zhang2023towards} introduce a Federated Instruction Tuning (FedIT) algorithm to leverage federated learning in induction tuning of FM to enhance their performance.
However, none of these works consider the cooperation of FMs and thus cannot achieve good performance in data heterogeneous conditions on challenging datasets.

\begin{figure*}[t]
  \centering
  \includegraphics[width=0.8\textwidth]{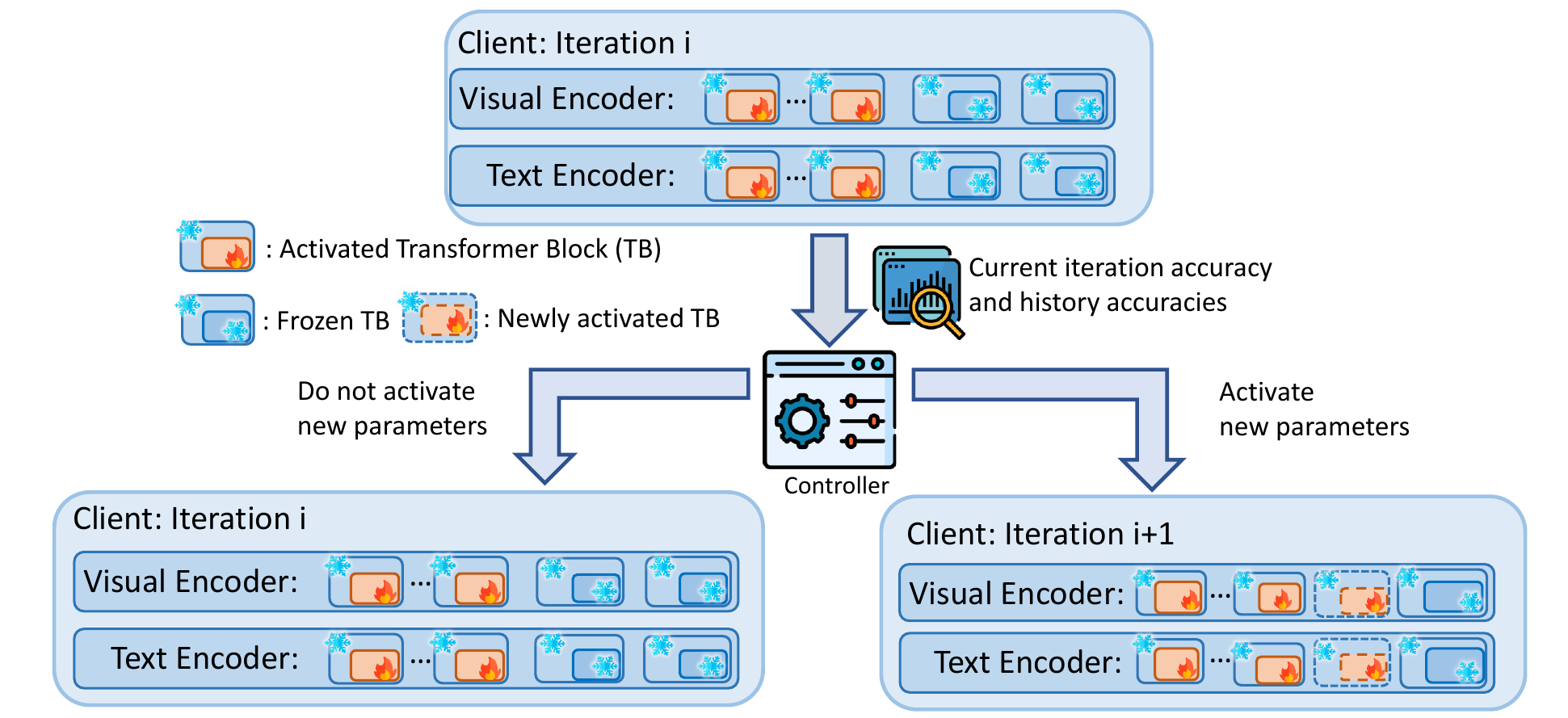}
  \vspace{-0.3cm}
  \caption{Iteration process of \texttt{SAL} }\label{SAL}
  \vspace{-0.3cm}
\end{figure*}

\section{Design of FedMS}

\subsection{Overview of FedMS}
We consider a typical FL scenario with a total number of $N$ clients with non-iid dataset $\{D_1, ..., D_N\}$. 
Our method FedMS consists of two stages of training as depicted in Fig.\ref{fig:FedMS}. 

In the first stage, low-rank adaptation matrices are inserted into every transformer block of the foundation model \cite{hu2021_lora}. All the clients freeze the pre-trained foundation model weights and only update and upload the weights of the inserted matrices in every communication round.
In this stage, every client collaboratively trains a global model $W_g$, which will be the global expert in stage two. The objective function of stage one can be expressed as
\begin{equation}
    \mathcal{F} = \frac{1}{N} \sum_{i=1}^N \mathbb{E}_{(\boldsymbol{x}_i, y_i) \sim {d}_i} \mathcal{L}_i(\boldsymbol{x}_i, y_i; W_g)
\end{equation}
where $\mathcal{L}_i$ is the loss function of client $i\in [N]$, $\boldsymbol{x}_i$ is its private data and ${y}_i$ is the corresponding label. $d_i$ denotes the data distribution of client $i$.

In the second stage, each client utilizes the trained global expert in the first stage and trains its personalized model. This model consists of a global expert, a local expert, and a gate model together constitute a Mixture of Foundation Models. During the second stage, local experts are only trained on clients' local datasets using a novel Sparsely Activated LoRA algorithm and do not engage the global aggregation. We optimize 
\begin{equation}
\mathcal{F}_i = \mathbb{E}_{(\boldsymbol{x}_i, y_i) \sim {d}_i} \mathcal{L}_i(\boldsymbol{x}_i, y_i; W_i)
\end{equation}
where $W_i$ is the parameters of the local expert of client $i$.

We design and insert a gate adapter into the gate model and aggregate all the parameters of gate adapters in each communication round. We optimize the gate adapter parameters by
\begin{equation}
    \mathcal{F}_{gate} = \frac{1}{N} \sum_{i=1}^N \mathbb{E}_{(\boldsymbol{x}_i, y_i) \sim {d}_i} \mathcal{L}_i(\boldsymbol{x}_i, y_i; G_i)
\end{equation}
where $G_i$ is the parameters of the gate model of client $i$. 

We propose two novel algorithms to tackle challenges raised by FL with FM.  


\subsection{Sparsely Activated LoRA}
According to \cite{kaplan2020scaling}, the capability of the deep neural network tends to improve with the increase of the number of parameters of the model. FL with FM presents substantial challenges to the communication and computation of the distributed system. 

In traditional FL \cite{mcmahan2017communication, li2020federated}, model parameters after local training are usually transmitted to the server for model weights aggregation in each communication round. 
This paradigm faces great challenges when FMs are trained in a FL procedure. 
Suppose we have a FM whose parameters are represented by $\mathbf{W_f}$. For full parameters fine-tuning, we need to calculate and store another model $\mathbf{W_k}$ which has the same parameter size of $\mathbf{W_f}$ for each task $k$. 
FM typically consists of over 10 million model parameters, resulting in significant transmission time requirements for modern mobile communication networks. Moreover, the training of FM necessitates substantial computation power and storage capacity, whereas edge devices typically possess limited computational capabilities and storage space. Therefore, it is imperative to develop an algorithm that mitigates the communication and computation costs associated with FL using FM.

To tackle these challenges, we design a novel Sparsely Activated LoRA (\texttt{SAL}) algorithm that can achieve the SOTA performance while only tuning less than $1\%$ of the total parameters of FM. 
Common pre-trained language models are capable of efficient learning even when randomly projected into a smaller subspace because they have a very low intrinsic dimension\cite{aghajanyan2021intrinsic}. Edward J. Hu et al.\cite{hu2021_lora} propose Low-rank adaptation (LoRA) to insert trainable low-rank decomposition matrices in FMs, enabling model optimization with minimal parameter tuning.

Inspired by this, we insert trainable low-rank decomposition matrices in every layer of the visual encoder and the text encoder of the CLIP model. We denote the weight parameter matrix as $W_0\in R^{E\times F}$ and the inserted low-rank decomposition matrices as $\Delta W$, which can be calculated by two low-rank matrices $\Delta W=W_A W_B$, $W_A\in R^{E\times H}$ and $W_B\in R^{H\times F}$ ($H << min(E, F)$). For $W_A$, we employ a random Gaussian initialization, while $W_B$ is initialized with zero. During training, $W_0$ is frozen and only $W_A$ and $W_B$ are optimized to save computation and storage costs.

Suppose the input of the weight matrices and the inserted low-rank decomposition matrices is $x$. The output can be calculated by 
\begin{equation}
 y = (W_0 + W_A W_B)x.
\end{equation}

The procedure of the proposed \texttt{SAL} algorithm is depicted in Fig.\ref{SAL}. We activate the low-rank decomposition matrices sparsely instead of activating them all during the training. At the beginning of the training stage, every layer of the visual encoder and the text encoder are inserted with frozen low-rank decomposition matrices. In deep neural networks, lower layers can better extract general information than higher layers\cite{yosinski2014transferable}. During the first training stage, low-rank decomposition matrices in all layers are activated to better extract general information to form a global expert while in the second stage, we unfreeze the low-rank decomposition matrices from higher layers to lower layers during the training. 

More specifically, we introduce a Capability Queue with a maximum queue length of $Q$. Image classification accuracies of clients are forwarded to the Capability Queue after every communication round. Once the Capability Queue is full, the previously added accuracies will be popped out. We set an accuracy threshold $\delta$ to help decide whether the training comes into a bottleneck. The incremental factor $\Delta$ of client $j$ in communication round $i$ is
\begin{equation}
    \Delta_{i,j} = Acc_{i,j} - \frac{1}{Q} \sum_{t=i-Q}^{i-1} Acc_{t,j}
\end{equation}
Where $Acc_{i,j}$ denotes the image classification accuracy of the model of client $j$ in communication round $i$. 
If $\Delta_{i,j} < \delta$, the training is considered to come into a bottleneck. Then low-rank decomposition matrices in the next lower layer will be activated. 

The design of the \texttt{SAL} algorithm is inspired by the fact that the performance of FM is usually affected by the model size, dataset size, and the quality of the dataset. Challenging datasets require more model parameters to be optimized to better extract the semantic meaning of the data. However, there is no silver bullet configuration in the training of FL with FM. So we introduce a Capability Queue to intelligently decide the number of tuning parameters and enable the training on computation resource-limited devices.



\begin{figure*}[htbp]
  \centering
  \begin{subfigure}[h]{0.32\textwidth}
    \includegraphics[width=\textwidth]{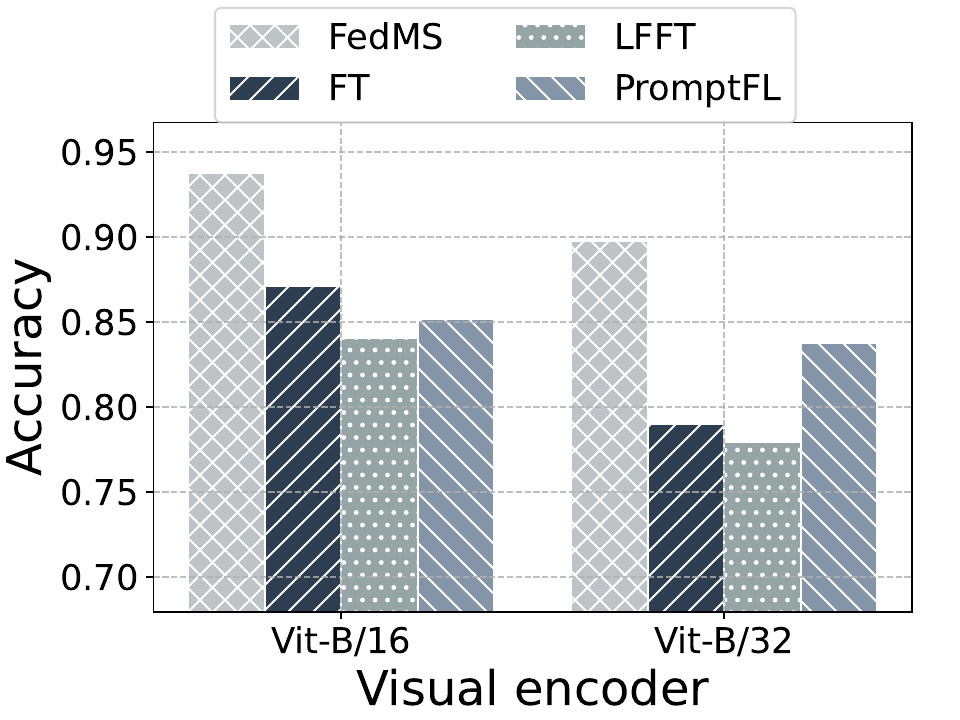}
    \caption{Food101}
    \label{fig:vitb32_1}
  \end{subfigure}
  \hfill
  \begin{subfigure}[h]{0.32\textwidth}
    \includegraphics[width=\textwidth]{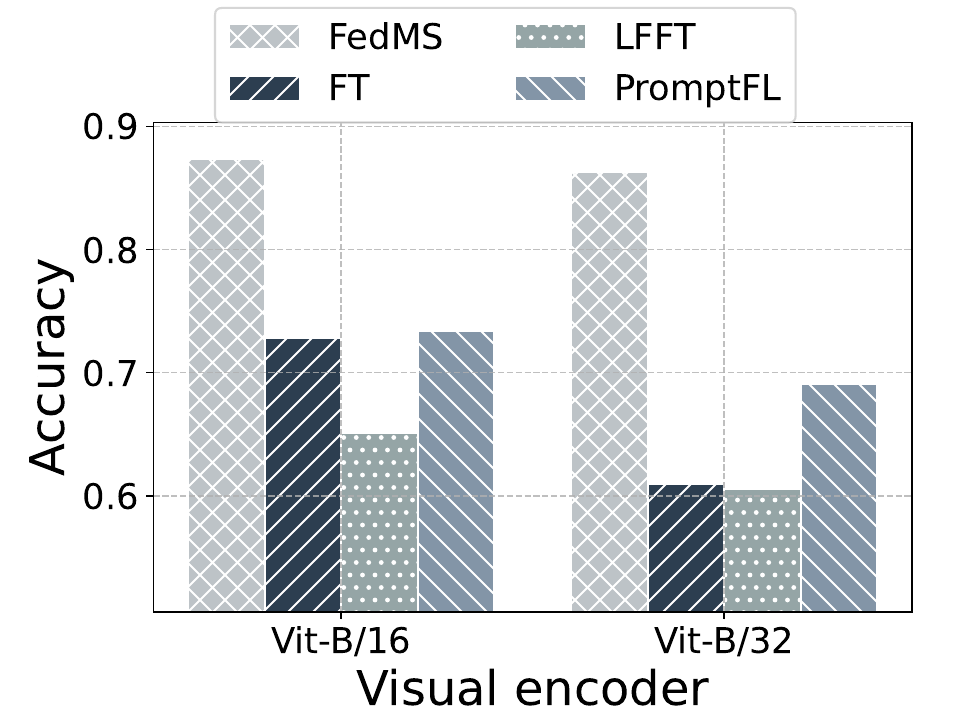}
    \caption{UCF101}
    \label{fig:vitb32_2}
  \end{subfigure}
  \hfill
  \begin{subfigure}[h]{0.32\textwidth}
    \includegraphics[width=\textwidth]{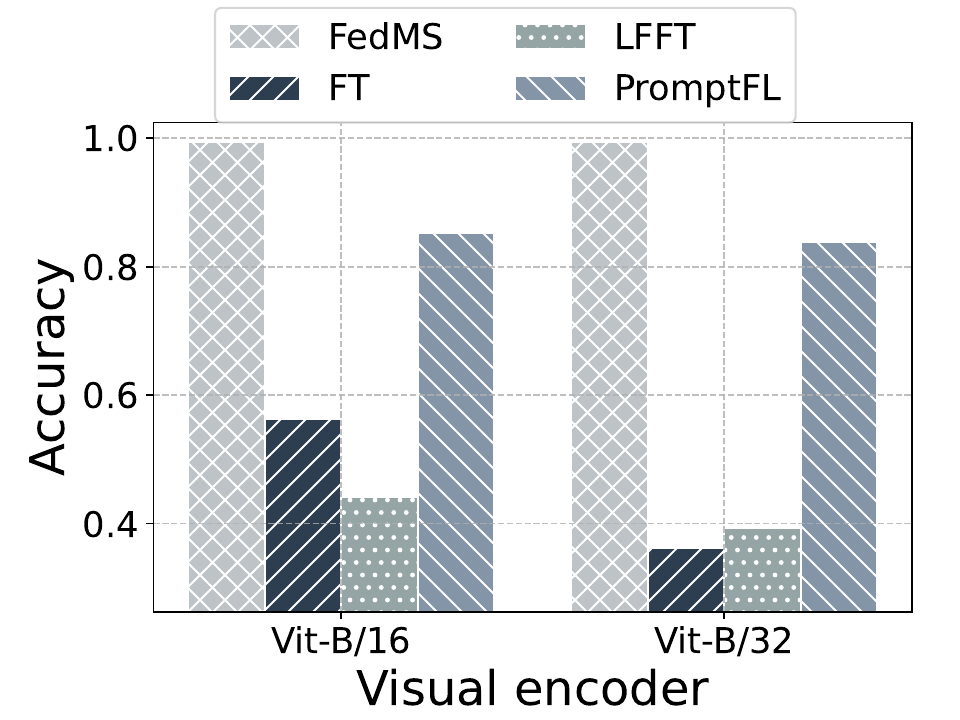}
    \caption{EuroSAT}
    \label{fig:vitb32_4}
  \end{subfigure}
  \caption{Average accuracy on different datasets when under different visual encoders}
  \label{fig:vitb32}
\end{figure*}


\subsection{Mixture of Foundation Models}
In traditional FL, a global model is trained using the decentralized data of clients. Only model weights are aggregated in the central server while the local data of clients are kept private to ensure clients' data privacy. This paradigm faces statistical challenges especially when the data distribution of clients is non-iid.  Such non-iid data distribution could cause the weight divergence during the training\cite{zhao2018federated} and cause significant performance drops. Moreover, training a single global model and applying it to all clients can not suit different clients' needs when their data have different data distributions. Training personalized models while benefiting from utilizing a global model is essential to providing better performance for different clients.

To tackle this challenge, we design a novel Mixture of Foundation Models (\texttt{MoFM}) algorithm to utilize an FM as the global expert and another FM as the local expert thus creating a mixture of Foundation Models to simultaneously learn personalized feature information as well as global feature information on each client. 

As shown in Fig.\ref{fig:FedMS}, in the first stage of training, every client collaboratively trains a global FM $\zeta_g$ with weight $W_g$. Low-rank decomposition matrices are inserted in every layer of the visual encoder and the text encoder. This global FM acts as a global expert. In the second stage, a local expert $\zeta_i$ with weight $W_i$ is created for each client $i$ to cooperate with the global expert. 

More specifically, the local experts have the same neural network architecture as the global expert and are initialized with the weights of the global expert. A gate function $G_i$ with weight $\xi_i$ for each client $i$ is a neural network introduced to control the relative contribution of the global expert and the local expert to the final image classification decision given different images. We denote the extracted image features and text features by the global expert as $V_g$ and $T_g$ and the extracted image features and text features by the local expert $i$ as $V_i$ and $T_i$. The final cosine similarity of image features and  text features extracted from the dataset of client $i$ can be denoted by
\begin{equation}
    \tilde{O_i} = \lambda_i <V_g,T_g> + (1-\lambda_i) <V_i, T_i>
\end{equation}
where $\lambda_i \in (0,1)$ is a weight factor representing the mixing ratio of the global expert and the local expert of client $i$. Larger $\lambda_i$ indicates more global knowledge is used while smaller $\lambda_i$ indicates more personal knowledge is used.

During the second training stage of FedMS, the weights of the global expert are frozen, and the local expert $\zeta_i$ and the gate model are optimized only using the local data of client $i$. 
The adapter\cite{houlsby2019parameter} has been a popular parameter-efficient tuning method in FMs. It works by inserting very few layers into FMs and optimizing FM by only tuning the inserted very few parameters. 

We design a gate adapter to adapt to the local datasets while maintaining a low computation and communication cost.
In each communication round, clients' activated gate adapter parameters are aggregated to learn global feature information, thus maintaining a low computation and communication cost.

We denote the gate adapter of gate $i$ as $Z_i$ and the gate adapter after aggregation as $Z_g$. Specifically, we construct the gate adapter with a Multi-Layer Perceptron (MLP), a batch norm layer, an MLP, a batch norm layer, and finally a Softmax function to ensure the output is between $(0,1)$. The gate adapter aggregation procedure is denoted as:   

\begin{equation}
    Z_g = \frac{\mathcal{L}_{i,j}}{\sum_{j=0}^N \mathcal{L}_{i,j}} \sum_{i=1}^N Z_i
\end{equation}
where $\mathcal{L}_{i,j}$ denotes the loss on the dataset of client $j$ in communication round $i$.

\begin{figure*}[htbp]
  \centering
  \begin{subfigure}[h]{0.32\textwidth}
    \includegraphics[width=\textwidth]{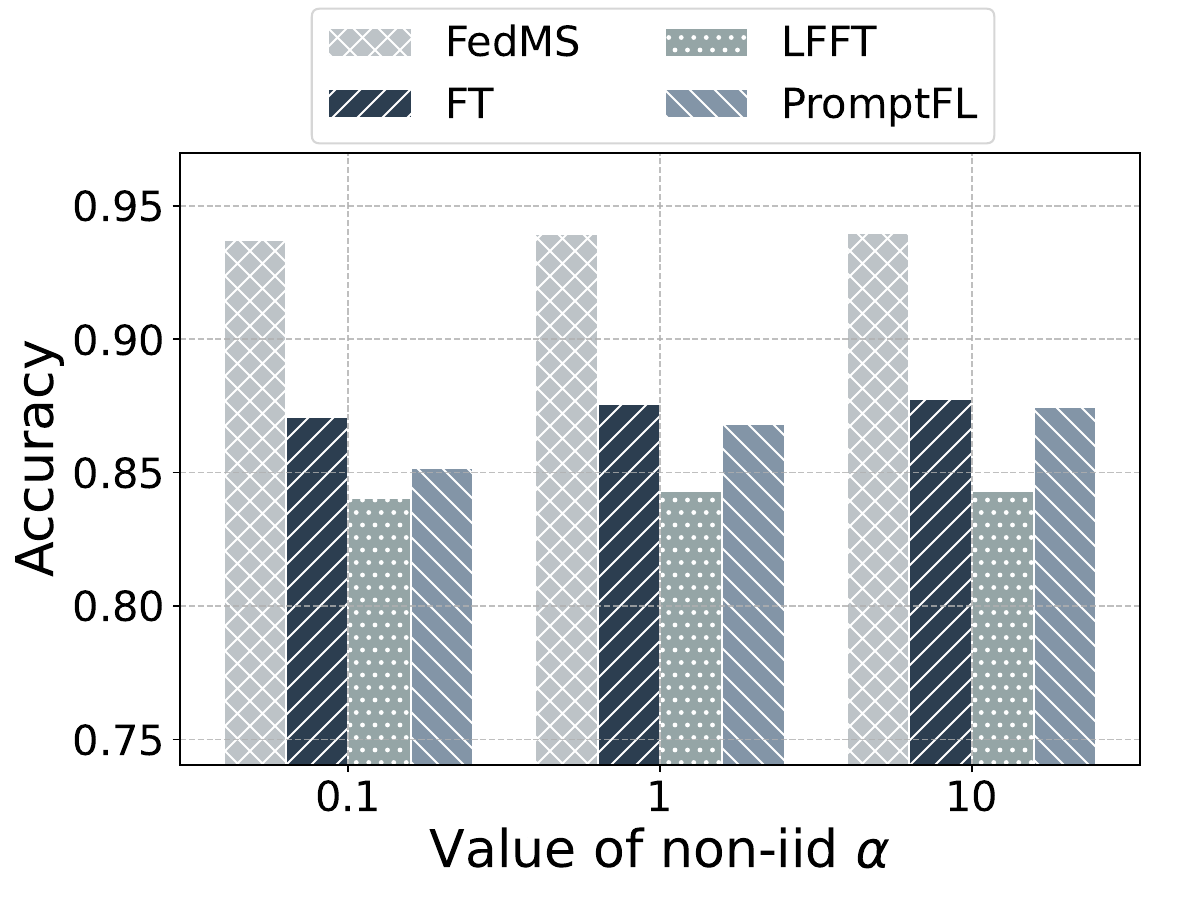}
    \caption{Food101}
    \label{fig:subfig1}
  \end{subfigure}
  \hfill
  \begin{subfigure}[h]{0.32\textwidth}
    \includegraphics[width=\textwidth]{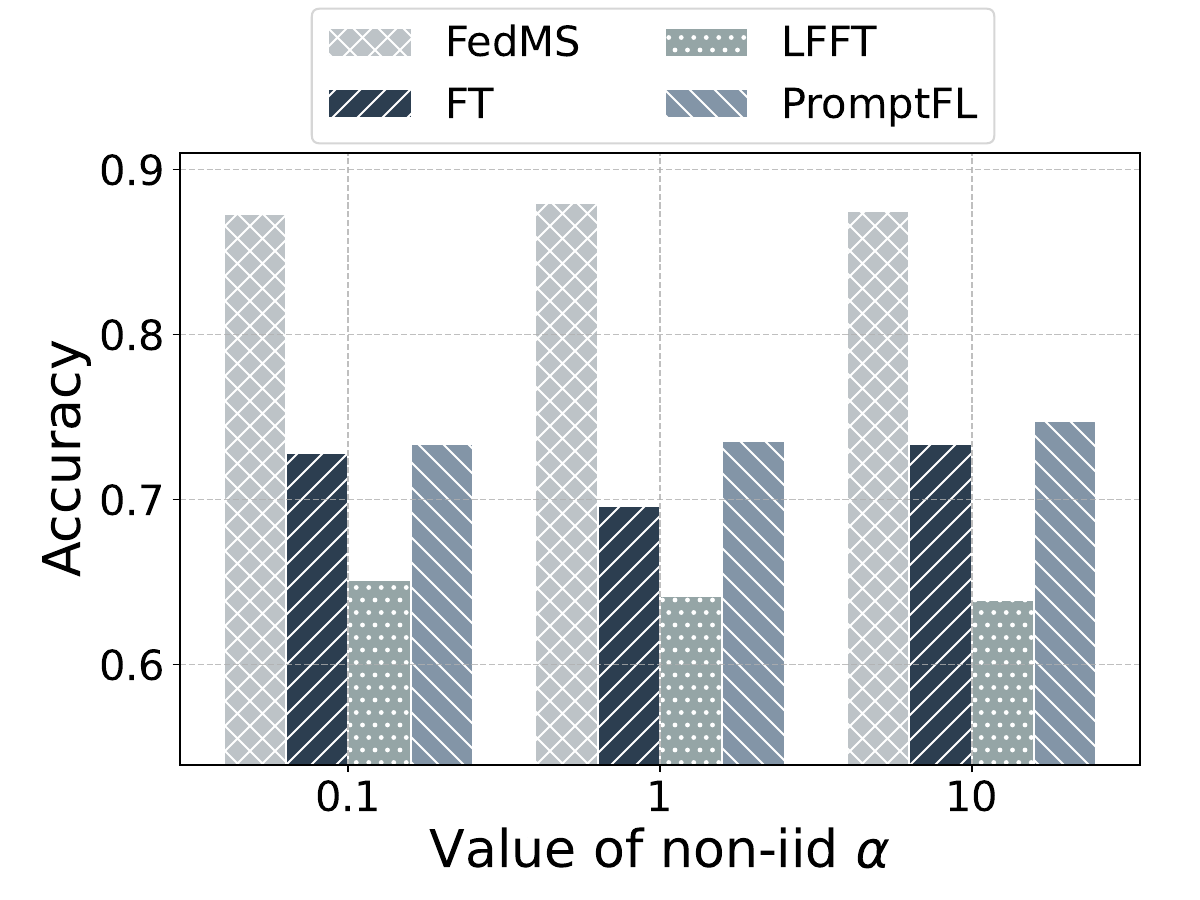}
    \caption{UCF101}
    \label{fig:subfig2}
  \end{subfigure}
  \hfill
  \begin{subfigure}[h]{0.32\textwidth}
    \includegraphics[width=\textwidth]{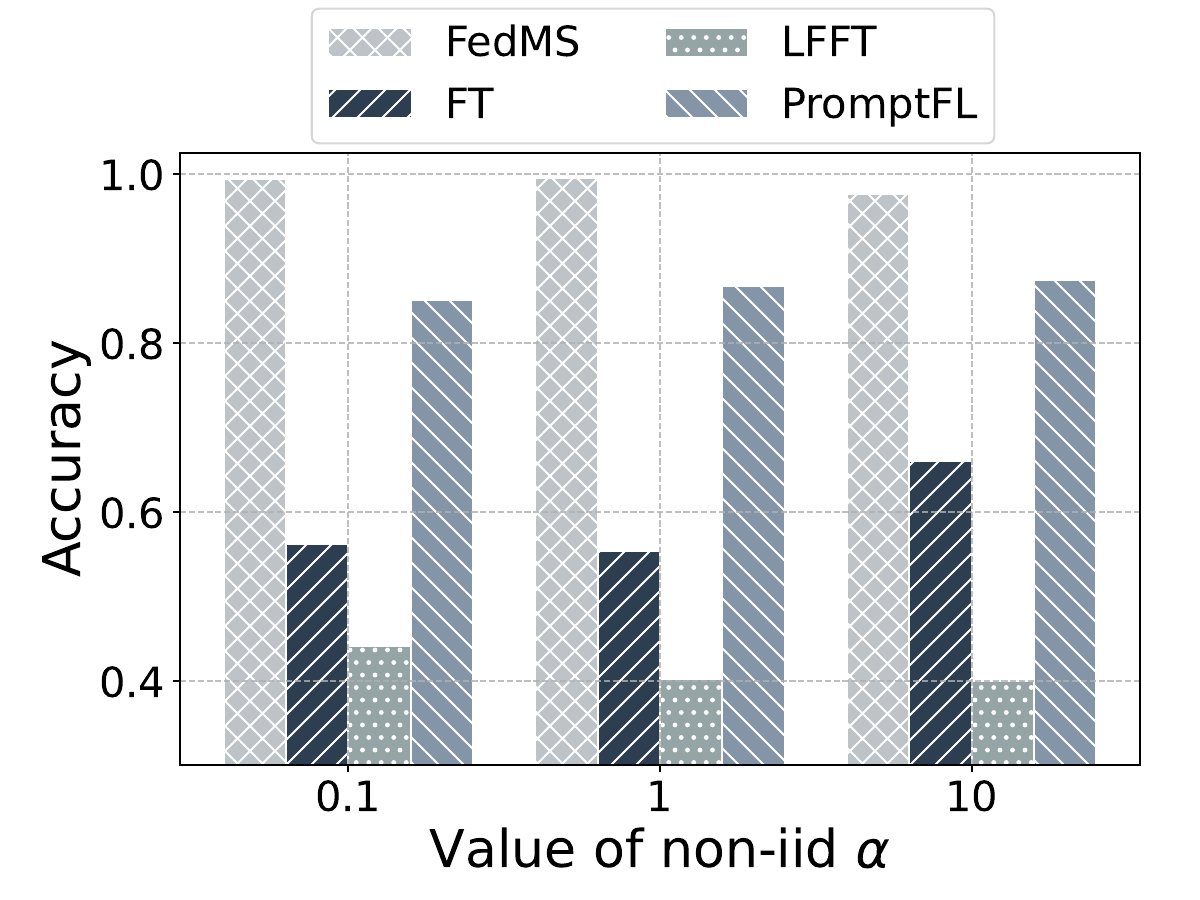}
    \caption{EuroSAT}
    \label{fig:subfig4}
  \end{subfigure}
  \caption{Average accuracy on different datasets at different non-iid $\alpha$}
  \label{fig:whole_figure_1}
\end{figure*}

\section{Experiments}
In this section, we conduct comprehensive experiments compared to SOTA baselines to verify the effectiveness of FedMS under different settings.
\subsection{Experiments set up}

\subsubsection{Datasets}
We select some representative datasets that are widely used in the image classification task of the CLIP model. Specifically, we select Food101\cite{bossard2014food} which is a food classification dataset containing 101 classes, EuroSAT\cite{helber2019eurosat} which is a dataset for land use and land cover classification containing 10 classes, and UCF101\cite{soomro2012ucf101} which is a dataset for human actions classification in the wild containing 101 classes.

\subsubsection{Baselines}
To verify the effectiveness of the proposed FedMS algorithm,  we compare image classification accuracy with the following state-of-the-art baselines.

\begin{itemize}
    \item \textit{Vanilla Fine-Tuning (FT)}: This is one of the most representative fin-tuning algorithms used in natural language processing and computer vision areas\cite{kenton2019bert}. 
    \item \textit{PromptFL\cite{guo2023promptfl}}: This algorithm does prompt tuning instead of tuning the parameters of FM. prompts from different clients are aggregated in every communication round.
    \item \textit{LayerFreeze Fine-Tuning (LFFT)}: This algorithm freezes several layers in FMs, and only the activated layers will be aggregated in every communication round to save communication and computation resources.
\end{itemize}

\begin{figure*}[htbp]
  \centering
  \begin{subfigure}[h]{0.32\textwidth}
    \includegraphics[width=\textwidth]{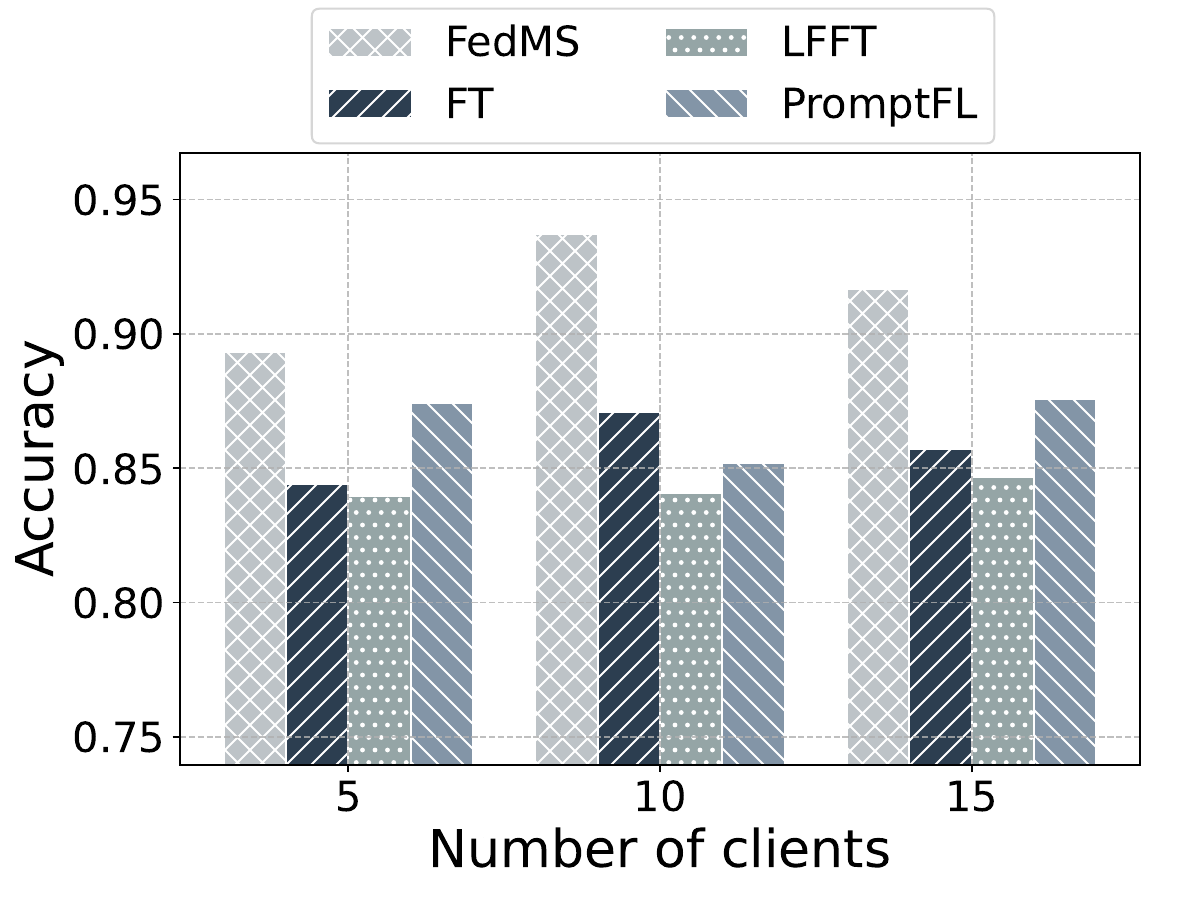}
    \caption{Food101}
    \label{fig:client5_1}
  \end{subfigure}
  \hfill
  \begin{subfigure}[h]{0.32\textwidth}
    \includegraphics[width=\textwidth]{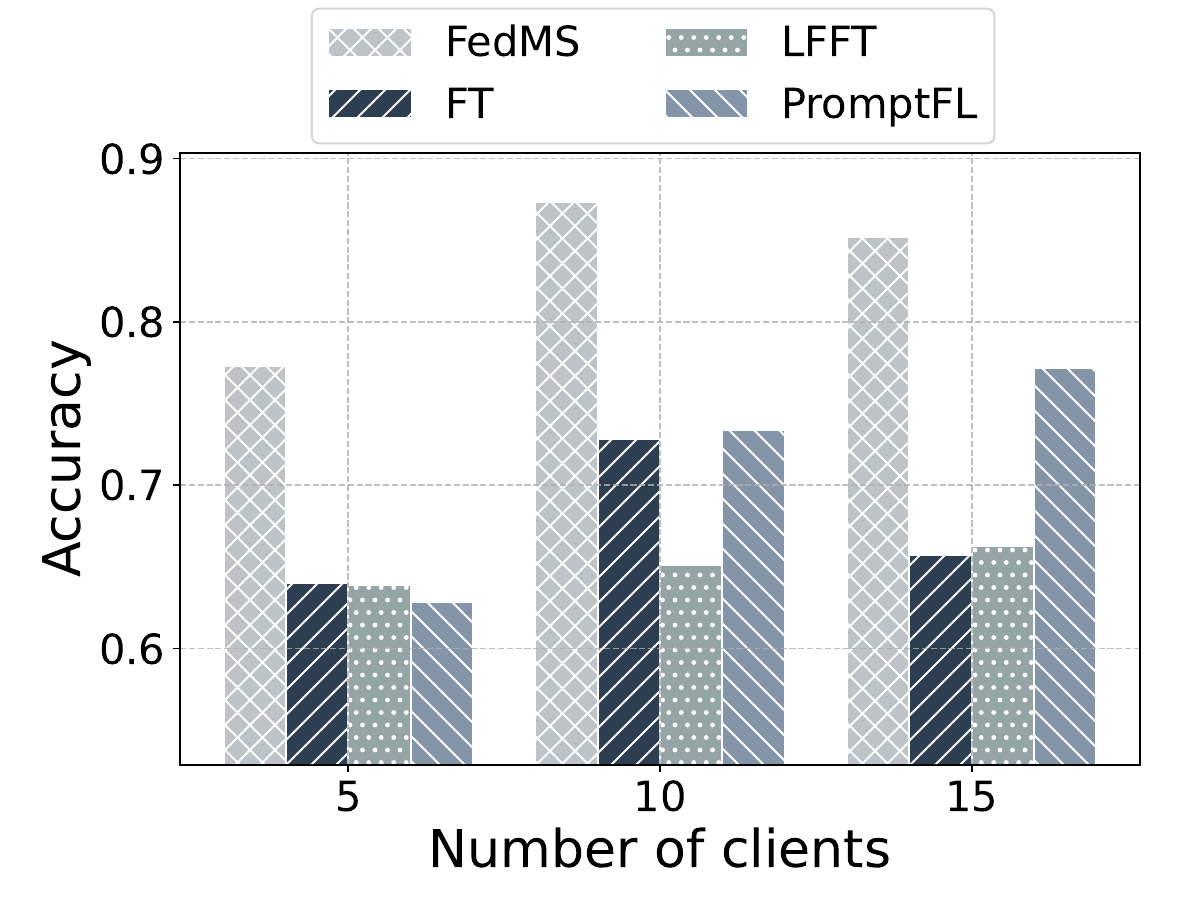}
    \caption{UCF101}
    \label{fig:client5_2}
  \end{subfigure}
  \hfill
  \begin{subfigure}[h]{0.32\textwidth}
    \includegraphics[width=\textwidth]{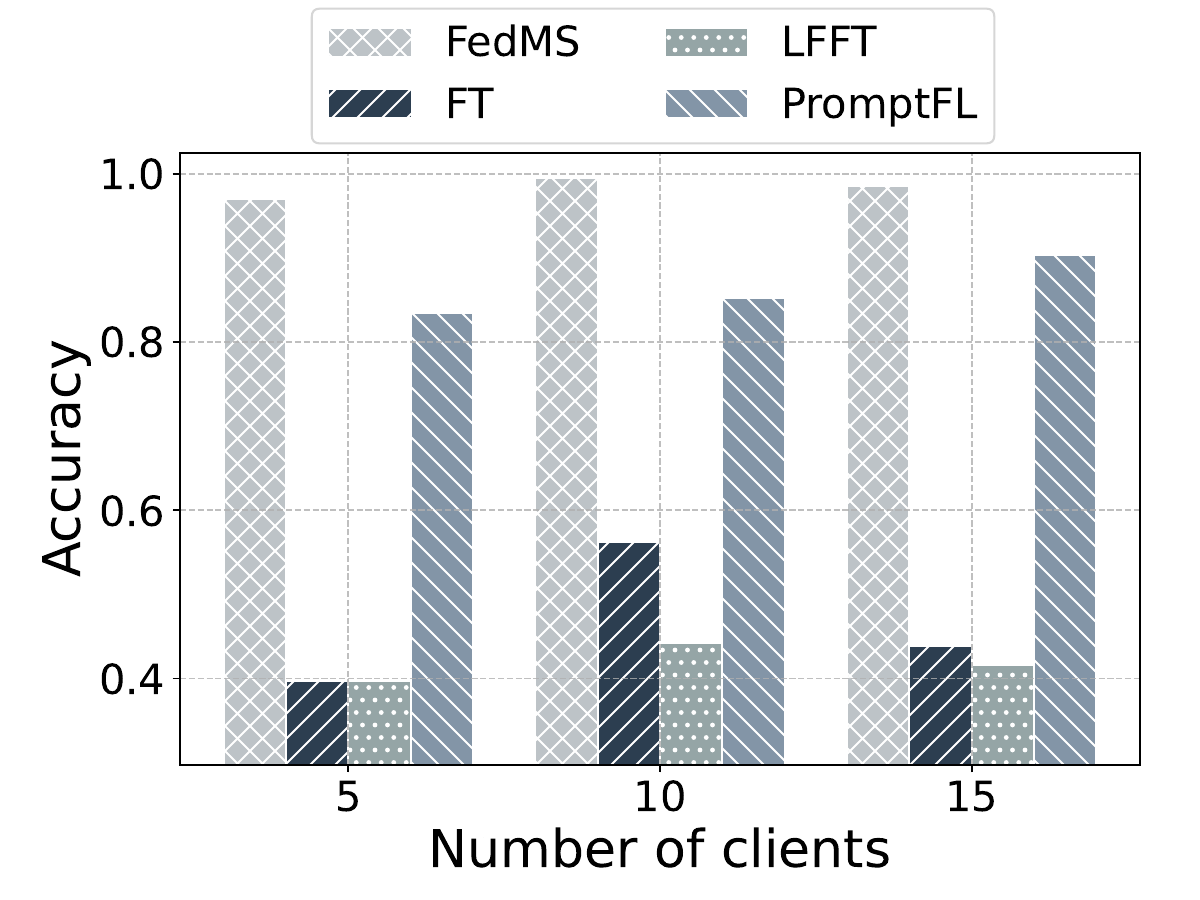}
    \caption{EuroSAT}
    \label{fig:client5_4}
  \end{subfigure}
  \caption{Average accuracy on different datasets under different number of clients}
  \label{fig:client5}
\end{figure*}

\begin{figure*}[h]
\centering
\hspace{0.1 cm}
\begin{minipage}{0.32\textwidth}
\centering
\includegraphics[width=\textwidth]{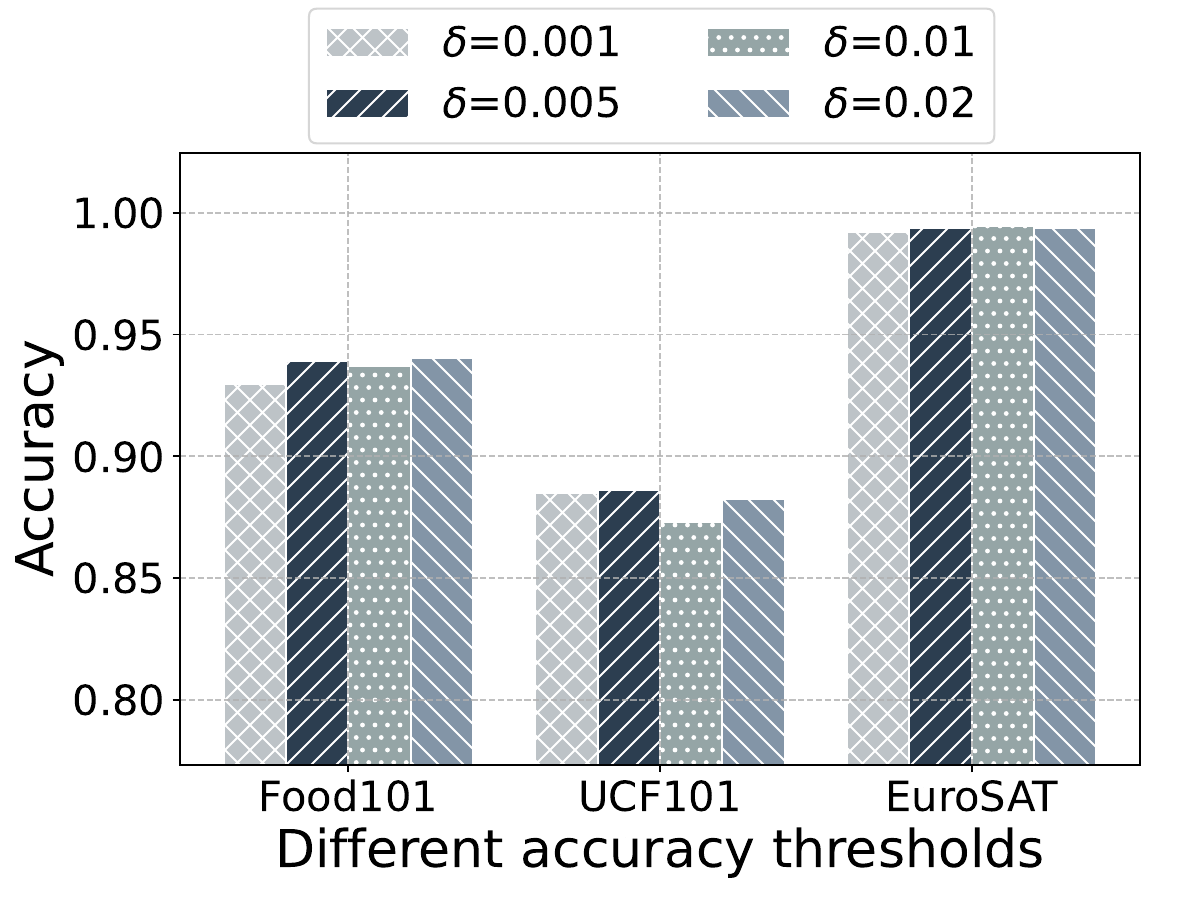}
\caption{Accuracy under different accuracy thresholds in capability queue}
\label{fig:diff_delta}
\end{minipage}%
\begin{minipage}{0.32\textwidth}
\centering
\includegraphics[width=\textwidth]{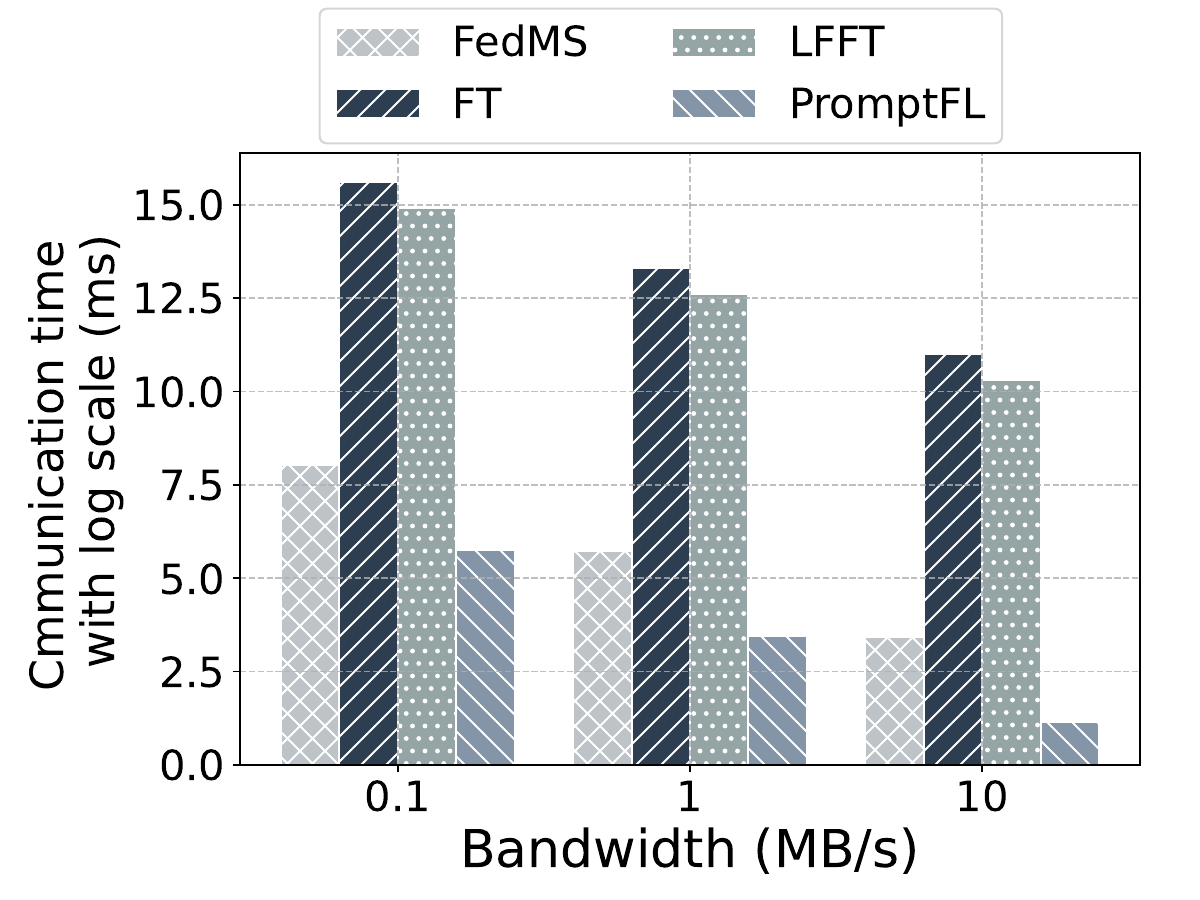}
\caption{Total communication time at different bandwidths}
\label{fig:diff_bandwidth}
\end{minipage}
\hspace{0.1 cm}
\begin{minipage}{0.32\textwidth}
\includegraphics[width=\textwidth]{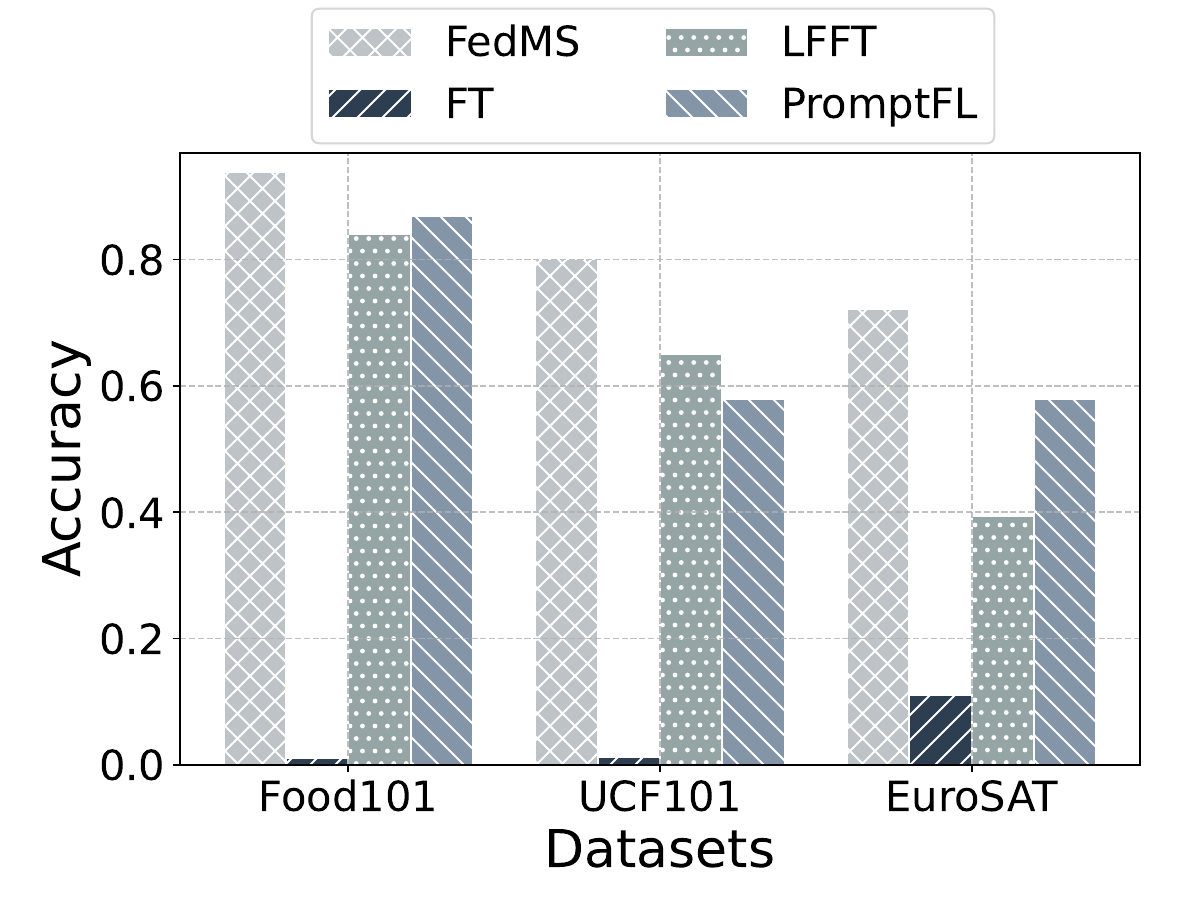}
\caption{Accuracy under backdoor attacks}
\label{fig:attack}
\end{minipage}
\hspace{0.1 cm}
\end{figure*}

\subsubsection{Default training settings}
We set the backbone of the visual encoder of CLIP to be ViT-B/16. The batch size is set to be 512. The learning rate is set to be $2 e^{-4}$. The optimizer is set to be Adam, with $\beta_1=0.9$, $\beta_2=0.98$, $\epsilon=1e^{-6}$, and weight decay to be 0.05. The number of clients is set to be 10. The rank of the inserted low-rank decomposition matrices is set to 1, 
The dropout probability of the inserted low-rank decomposition matrices is set to 0.1. The number of communication rounds of training stage one is set to 25 and the number of communication rounds of training stage two is set to 25.

\subsection{Results Comparisons}


We assume the non-iid data partition in FL to follow the Dirichlet distribution \cite{yurochkin2019bayesian}. The $\alpha$ parameter in Dirichlet distribution represents the degree of heterogeneity. The smaller the $\alpha$, the more non-iid the data distributed in the clients will be.
We test the image classification accuracy of various datasets in different non-iid levels.

\subsubsection{Impact of different system settings}

\textbf{Impact of degrees of data heterogeneity.}
Fig. \ref{fig:whole_figure_1} shows the image classification accuracy of Food101, UCF101, and EuroSAT datasets respectively, assuming different degrees of data heterogeneity. Specifically, we set the $\alpha$ to be 0.1, 1, and 10. From the results, we can find that FedMS achieves the highest accuracy among the four algorithms in all cases. FedMS has surpassed the average accuracy of FT, PromptFL, and LFFT in all datasets by  $20.57\%, 30.06\%$, and $11.17\%$ respectively. Our method has a minimum accuracy gain of $6.36\%, 14.43\%$, and $12.71\%$ and a maximum accuracy gain of $9.62\%, 23.82\%$, and $59.21\%$ on Food101, UCF101, and EuroSAT datasets respectively compared to other baselines when $\alpha$ is set to 1. 

By observing the accuracy at different data heterogeneity levels, we can find that the performance of these algorithms does not always follow a positive correlation relationship with the data heterogeneity level. On Food101 and UCF101 datasets, the accuracy of FedMS when $\alpha$ is 10 has an increase of $0.24\%$ and $0.15\%$ compared to the case when $\alpha$ is 0.1. On the EuroSAT dataset, FedMS has an accuracy of $99.46\%$ when $\alpha=1$, but has an accuracy of $97.67\%$ when $\alpha=10$.

This is because higher data heterogeneity may lead to a smaller number of classes in clients' local datasets. For example, a client's local dataset may contain data from 10 classes at a low data heterogeneity level but may contain 3 classes at a high data heterogeneity level, which can lower the difficulty of identifying the right class given an image.

\textbf{Impact of number of clients.}
We test the performance of FedMS and other baselines under different number of clients. Specifically, we set the number of clients to 5, 10, and 15.

From Fig. \ref{fig:client5} we can conclude that FedMS has the highest accuracy in different client number cases. 
When the number of clients is 5, FedMS achieves the accuracy of $91.67\%, 85.18\%$, and $98.51\%$ in Food101, UCF101, and EuroSAT datasets respectively while the best accuracy of the other three baselines in these three datasets are $87.56\%, 77.15\%,$ and $90.40\%$. FedMS suppresses the best performance of other baselines by $4.11\%, 8.03\%$, and $8.11\%$. 
In the case when there are 10 clients, FedMS has a maximum accuracy increase of $9.67\%, 22.2\%$, and $55.24\%$ compared to the three baselines.
When the number of clients reaches 15, the accuracy of FedMS is $89.31\%, 77.31\%$, and $97.03\%$ while the highest accuracy of the other three baselines are 
$87.44\%, 62.86\%$, and $83.44\%$. Results show that FedMS works well under different scales.

\begin{figure*}[h]
\centering
\begin{minipage}{0.32\textwidth}
\centering
\includegraphics[width=\textwidth]{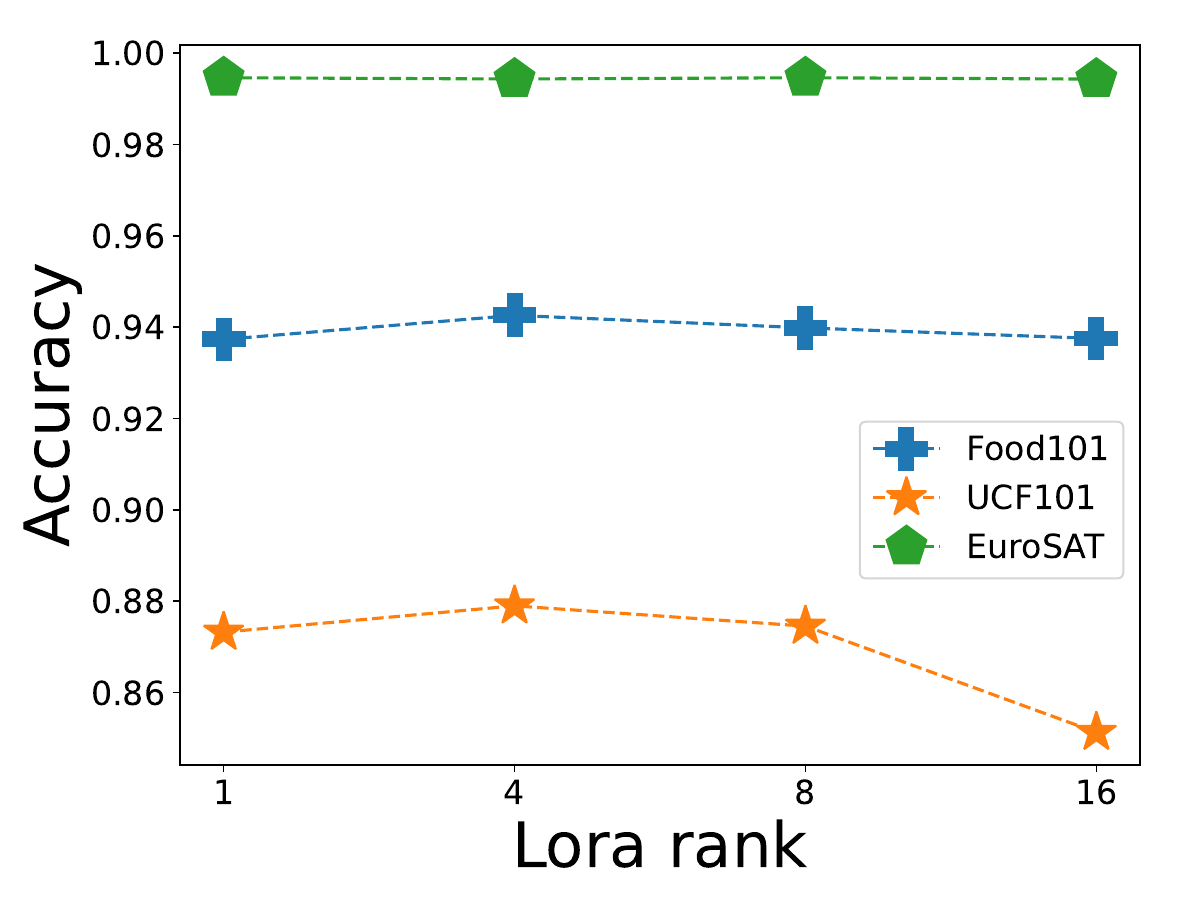}
\caption{Accuracy with different LoRA ranks}
\label{fig:LoRA_rank}
\end{minipage}%
\hspace{0.1 cm}
\begin{minipage}{0.32\textwidth}
\centering
\includegraphics[width=\textwidth]{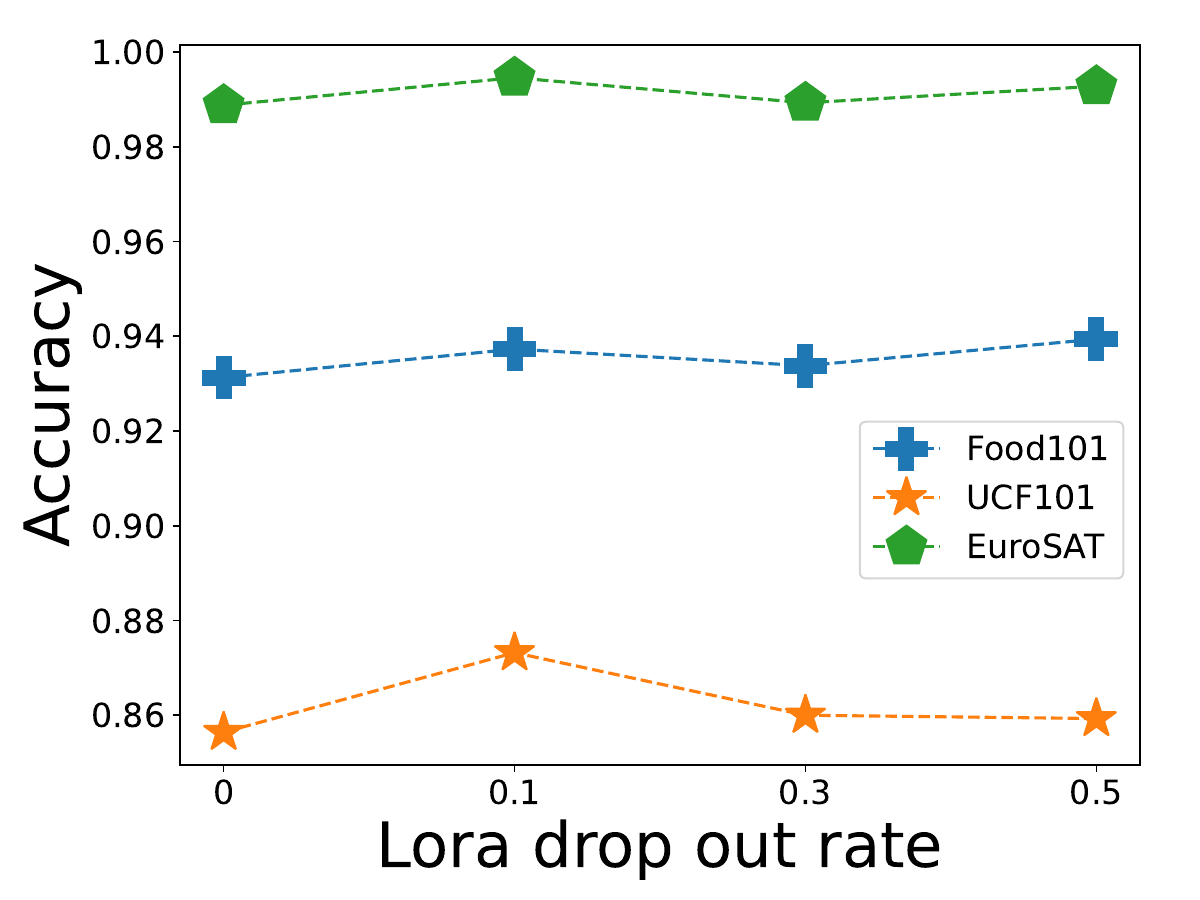}
\caption{Accuracy with different LoRA dropout rates}
\label{fig:LoRA_dropout}
\end{minipage}%
\hspace{0.1 cm}
\begin{minipage}{0.32\textwidth}
\includegraphics[width=\textwidth]{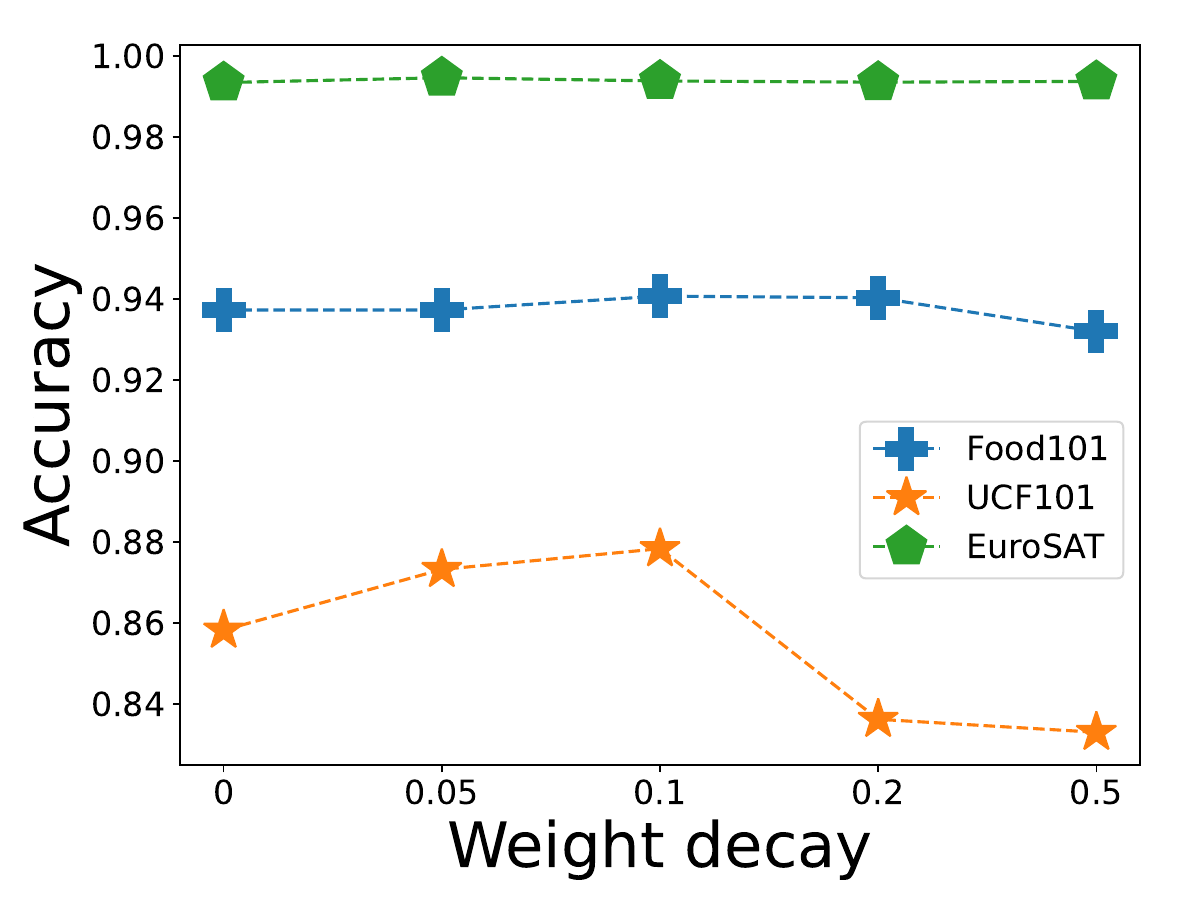}
\caption{Accuracy with different LoRA weight decays}
\label{fig:LoRA_decay}
\end{minipage}
\vspace{-0.3cm}
\end{figure*}

\begin{figure}
  \centering
  \begin{subfigure}[b]{0.5\textwidth}
    \includegraphics[width=\textwidth]{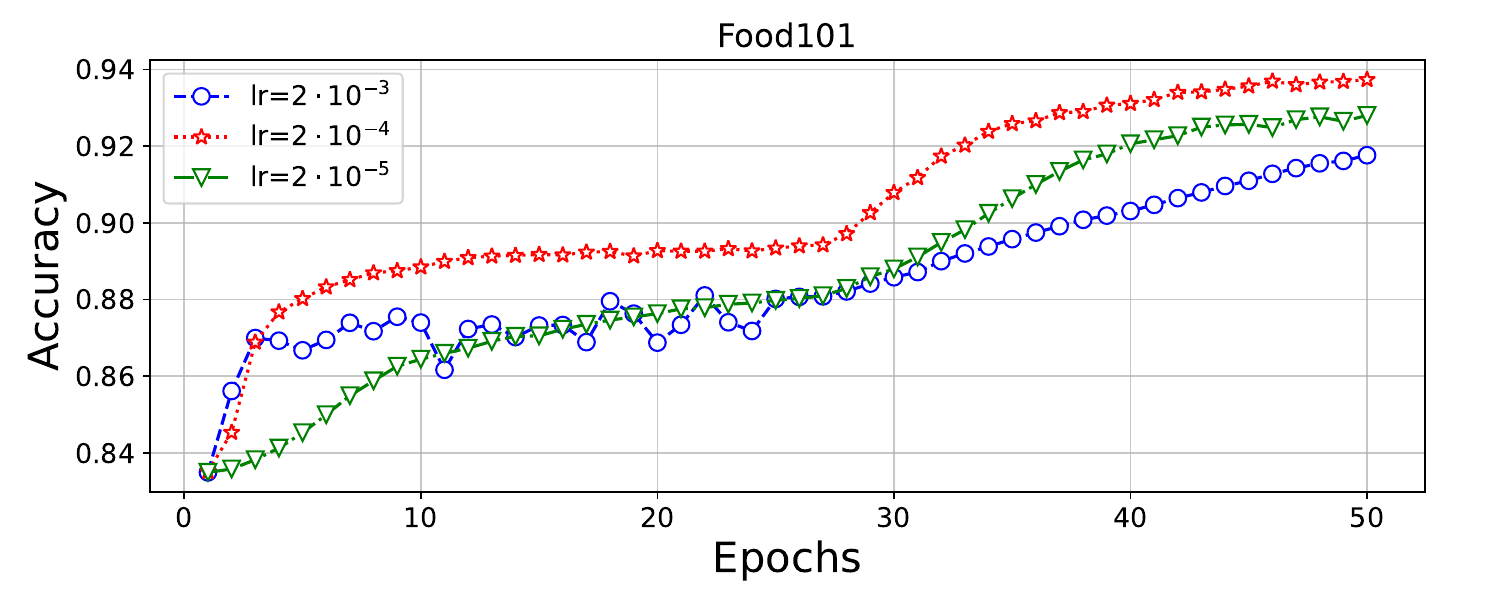}
    \label{fig:subfig1}
  \end{subfigure}
    \vspace{-15pt} 
  \begin{subfigure}[b]{0.5\textwidth}
    \includegraphics[width=\textwidth]{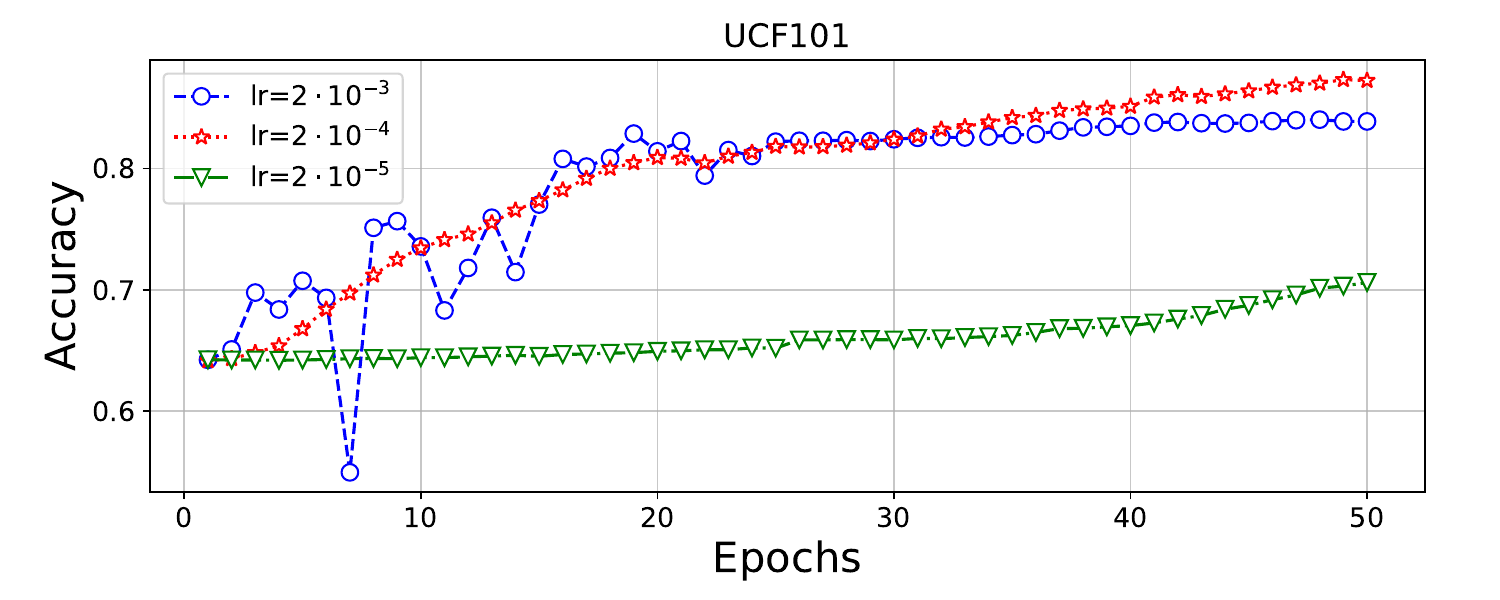}
    \label{fig:subfig2}
  \end{subfigure}
    \vspace{-10pt} 
  \begin{subfigure}[b]{0.5\textwidth}
    \includegraphics[width=\textwidth]{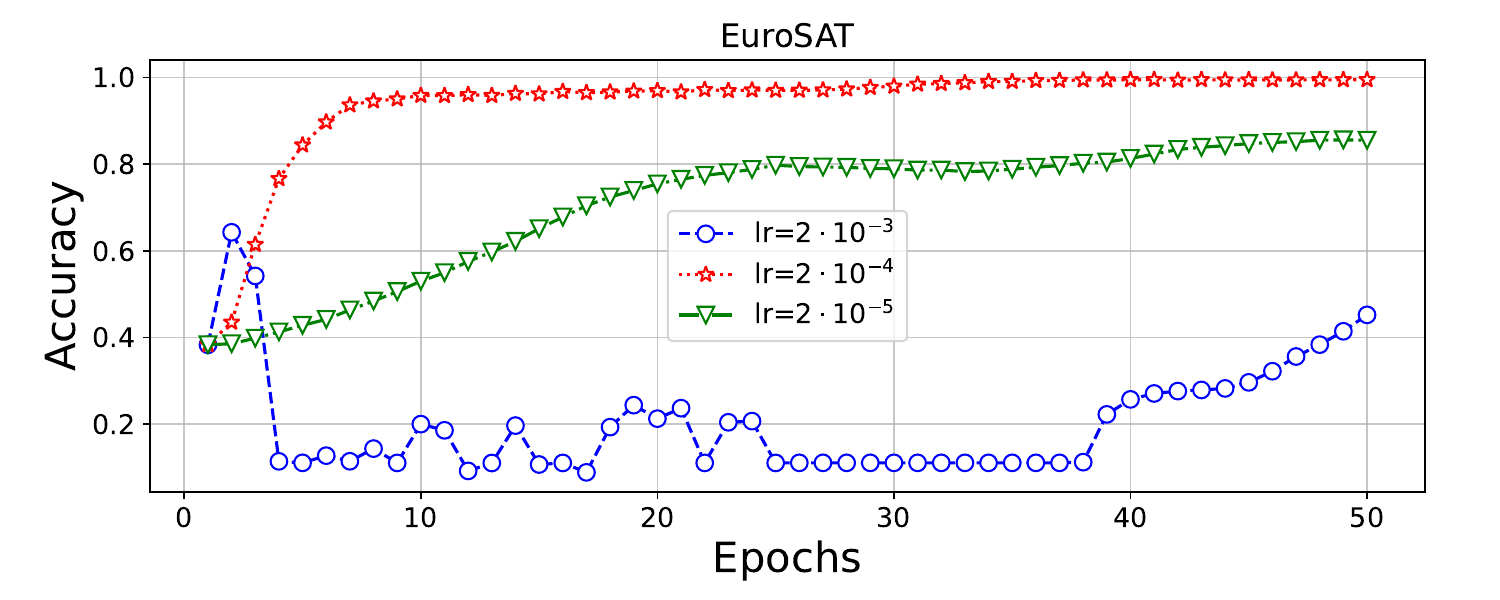}
    \label{fig:subfig3}
  \end{subfigure}
    \vspace{-10pt}
  \caption{Comparison of accuracy of FedMS under different learning rates}
  \label{fig:diff_lr}
\end{figure}

\subsubsection{Impact of training settings}

\textbf{Impact of visual encoders.}
We test the performance of FedMS under visual encoders ViT-B/16 and ViT-B/32 to further verify the effectiveness of FedMS under various visual encoders.

We can observe from Fig. \ref{fig:vitb32} that FedMS achieves the highest image classification accuracy on all datasets using the visual backbone Vit-B/16 or Vit-B/32. Results show that visual encoders with a larger number of parameters can achieve better performance than those with smaller visual encoders. The accuracy of the four algorithms increased by $1.01\%, 11.85\%, 4.57\%$, and $4.25\%$ when using ViT-B/16 as the visual encoder on the UCF101 dataset which confirms the theoretical analysis that larger models have better feature extraction ability.

Our method suppresses other baselines by a maximum of $55.25\%$ and a minimum of $8.55\%$ when using Vit-B/16 and suppresses other baselines by a maximum of $60.09\%$ and a minimum of $6.01\%$ when using Vit-B/32. Results show that FedMS can adapt to visual encoders with different scales.

\textbf{Impact of accuracy thresholds.}
We further discuss the performance of FedMS when the value of the accuracy threshold $\delta$ is different. Typically, if the clients have more computation resources, they can have a higher $\delta$ to encourage more inserted low-rank decomposition matrices to be activated, and if their computing resources are scarce they can set a small $\delta$ to save more computation resources.

We set the accuracy threshold to 0.001, 0.005, 0.01, and 0.02 to see its effect on the model performance.
From Fig. \ref{fig:diff_delta} we can find that the accuracy does not always increase with $\delta$. On the Food101 dataset, the accuracy is $94.05\%$ when $\alpha$ is 0.02 and is $92.96\%$ when $\alpha$ is 0.001, which has an increase of $1.09\%$ and on the EuroSAT dataset, the accuracy increase from $99.23\%$ to $99.39\%$. But on the UCF101 dataset, the accuracy decreased from $88.48\%$ to $88.24\%$ when $\delta$ increased from 0.001 to 0.02. The largest accuracy difference in the three datasets is $1.09\%, 0.0024\%$, and $0.0016\%$. 

Results imply the effectiveness of the design of \texttt{SAL} because of leveraging the idea of curriculum learning that progressively increases the number of activates low-rank adaptation matrices in the visual encoder and the text encoder. 
In the optimization process of FM, it is often easier at the beginning, while the optimization of parameters becomes more difficult as it progresses, making it harder to improve accuracy. We increase the number of activated parameters through a controller during the training to tackle the increasing difficulty of optimizing the FM during training.

\textbf{Imapct of learning rates.}
We show the accuracy of FedMS in three datasets under different learning rates. We set the learning rate to $2 e^{-3}, 2 e^{-4}$, and $2 e^{-5}$.

From Fig. \ref{fig:diff_lr} we can observe that when the learning rate is $2 e^{-5}$ the model shows a slow convergence rate, especially in UCF101 datasets, the accuracy is $70.61\%$ which has an accuracy loss of $16.72\%$ compared to the accuracy when the learning rate is $2e^{-4}$. When the learning rate is $2e^{-3}$, the accuracy has a sharp increase in the first few epochs in all datasets but may encounter severe oscillation in the following epochs which is because of the large learning rate can cause instability in model training. This phenomenon is especially usual in the training of FMs for the reason that they usually have a large number of parameters and the scale of gradient calculations will increase which may cause the gradient explosion. Moreover,  models with large parameters tend to have more complex optimization spaces, resulting in the training process being more easily affected by noise and instability.

\textbf{Impact of LoRA ranks.}
In FedMS, we incorporate LoRA for model training and optimization. The rank in LoRA refers to the degree of model compression or pruning, and different rank settings can have an impact on the model training performance. We examine the training accuracy performance of FedMS under different LoRA ranks, specifically rank is set to 1, 4, 8, and 16.

From Fig. \ref{fig:LoRA_rank}, we can observe that as the LoRA rank increases, FedMS shows a certain trend in accuracy performance across different datasets: it performs the best at LoRA rank=4, but the model accuracy decreases as the LoRA rank further increases. This trend is particularly evident on the UCF101 dataset, where changes in LoRA rank can cause the performance of the final model to fluctuate within a 5\% range. 
One possible reason is that the semantic information that the UCF101 dataset contains is difficult for the model to capture and thus requires more training to optimize the model. A high LoRA rank can lead to a large optimization space which raises challenges for the training. On the other hand, the information patterns in the Food101 and EuroSAT datasets are relatively simpler and easy for the inserted low-rank matrices with different ranks to capture, so the model's learning performance is not affected significantly. 


\textbf{Impact of LoRA dropout rates.}
The dropout coefficient in LoRA affects the probability of applying dropout regularization during the model training process. In FedMS, we apply dropout to prevent neural networks from overfitting and enhance the model's generalization ability. In our experiment, the group with a dropout rate of 0 will serve as the control group to represent the accuracy of the model without using dropout techniques. The groups with dropout rates of 0.1, 0.3, and 0.5 will serve as the experimental groups to investigate the impact of different dropout coefficients.

From Fig. \ref{fig:LoRA_dropout}, we can clearly see that dropout has a significant effect on the final model accuracy when set to 0.1. It results in a noticeable improvement of 0.5\%, 2\%, and 0.6\% respectively on the Food101, UFC101, and EuroSAT datasets. This improvement is quite significant, especially when the model itself already has a high accuracy on the dataset. Similar to the LoRA rank experiment, FedMS shows a decrease in accuracy on the UFC101 dataset with higher dropout rates. This is expected due to the more complex information patterns in the UFC101 dataset. Setting dropout too high increases the number of discarded neurons during model training, leading to a decrease in the model's learning and expressive capacity. Similarly, due to the simpler information patterns in the Food101 and EuroSAT datasets, the higher dropout rates contribute to the improved accuracy of the FedMS model on these two datasets.

\textbf{Impact of weight decays.}
Weight decay is used to solve the overfitting problem. By adding a regularization term to the loss function, it can encourage the model to have small weight values during the training. In our experiment, we set up five different experimental groups with varying degrees of weight decay. The group with a weight decay value of 0 represents the training performance of FedMS without using weight decay.

From Fig. \ref{fig:LoRA_decay} we can conclude that weight decay has a different impact on different datasets. On the UCF101 dataset, it has a significant impact on the accuracy. When the weight decay is 0.5 it has an accuracy loss of $4.53\%$ compared to the case when weight decay is 0.1. Weight decay has less impact on Food101 and UCF101 datasets, the accuracy under different weight decay varies less than $1\%$ on these two datasets. Overall, the performance of FedMS improves with the use of weight decay but decreases when the weight decay value is set too high. This is because large weight decay can lead to excessive constraints on parameters, limiting the effective information learned by the model.


\textbf{Impact of backdoor attacks.}
To further verify the robustness of FedMS we test the performance of FedMS and the three baselines under backdoor attack. We suppose that there is a certain ratio of malicious clients controlled by the attacker. The controlled malicious clients update the reverse weight of their local model in the weight aggregation every communication round to attack the FL system.

We set the ratio of malicious clients to the total number of clients to $20\%$ and compare the accuracies. From Fig. \ref{fig:attack} we can observe that the backdoor attack can severely harm the performance of the algorithm. FedMS has the highest accuracy in all datasets. It has an accuracy gain of  $6.97\%, 14.94\%$, and $14.23\%$ on three datasets compared to the highest accuracy of the three baselines. On the UCF101 dataset, the accuracy of the FT, LFFT, and, PromptFL have an accuracy of $1.20\%, 65.09\%$, and $57.90\%$ respectively. Three baselines have the lowest average on the EuroSAT dataset which is $36.11\%$, and they achieve the highest average accuracy of $86.88\%$ on the Food101 dataset. FMs are pre-trained on large-scale datasets which enables them to have strong zero-shot ability. The FM we use has the highest zero-shot accuracy on the Food101 dataset among the three datasets which makes it more resistant to backdoor attacks when it is trained on the Food101 dataset. The FT algorithm is the most easily poisoned because it exposes all parameters to the attack. Although the LFFT algorithm performs inferior to FT when there are no attacks, it outperforms FT under backdoor attacks due to the reason that the frozen transformer layers can help maintain the feature extraction ability of FM.




        

\begin{table}[]
\centering
\begin{tabular}{|c|c|}
\hline
Methods         & Proportion \\ \hline
FedMS (Stage 1) & 0.1\%      \\ \hline
FedMS (Stage 2)  & 0.27 \%    \\ \hline
FT             & 100\%      \\ \hline
LFFT            & 50\%       \\ \hline
PromptFL        & 0.005\%    \\ \hline
\end{tabular}
    \caption{Proportion of training parameters of different methods}
    \label{tab:parameter}
\end{table}

\begin{figure*}[htbp]
  \centering
  \begin{subfigure}[h]{0.32\textwidth}
    \includegraphics[width=\textwidth]{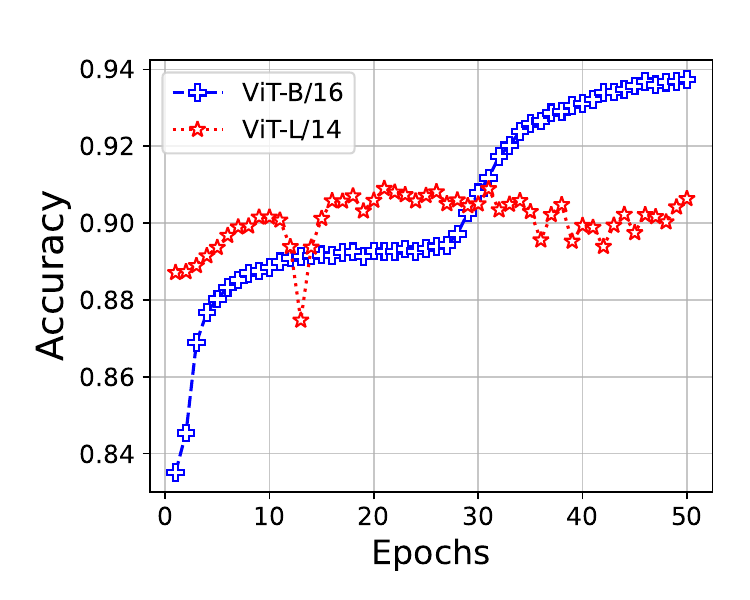}
    \caption{Food101}
    \label{fig:client5_1}
  \end{subfigure}
  \hfill
  \begin{subfigure}[h]{0.32\textwidth}
    \includegraphics[width=\textwidth]{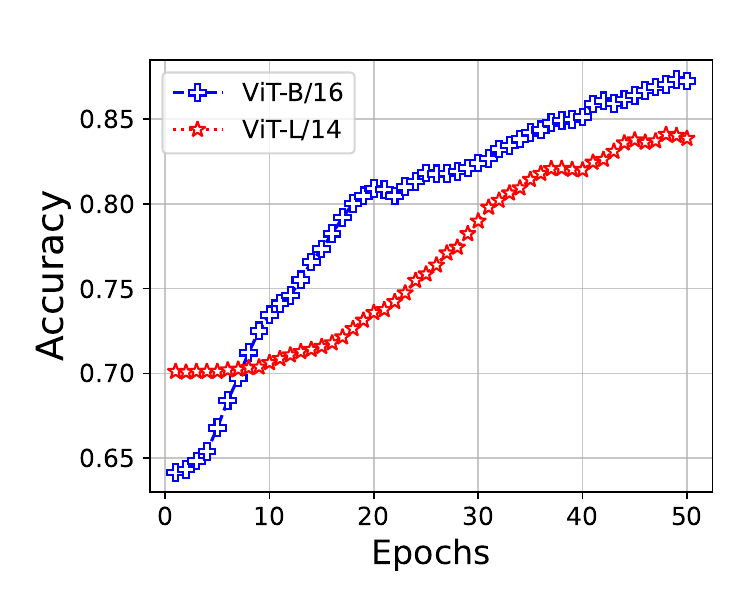}
    \caption{UCF101}
    \label{fig:client5_2}
  \end{subfigure}
  \hfill
  \begin{subfigure}[h]{0.32\textwidth}
    \includegraphics[width=\textwidth]{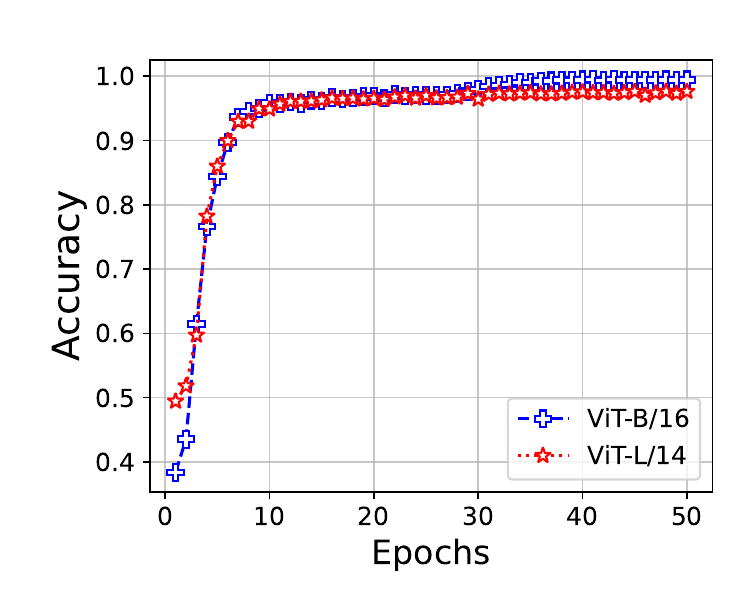}
    \caption{EuroSAT}
    \label{fig:client5_4}
  \end{subfigure}
  \caption{Comparison of the accuracy on different datasets using model with \texttt{MoFM} (ViT-B/16) and model without \texttt{MoFM} (ViT-L/14)}
  \label{fig:Comparison_diff_vit}
\end{figure*}

\subsubsection{Comparison of resource consumption}

\textbf{Comparison of the number of training parameters.}
Tab. \ref{tab:parameter} illustrates the proportion of the number of training parameters of FedMS and other baselines to the total parameters of the FM. Results show that FedMS only tunes $0.1\%$ and $0.108\%$ of parameters on training stage one and training stage two. FT and LFFT tune $100\%$ and $50\%$ of parameters. Notice that although PromptFL can save $0.095\%$ to $0.103\%$ trainable parameters compared to FedMS, it has a cost of severe performance drop. FedMS achieves the best trade-off between training parameter saving and model performance.

\textbf{Comparison of transmission time.}
Considering the scenario in mobile computing and the Internet of Things, we test the proposed FedMS in different bandwidth conditions. Specifically, we set the bandwidth to be 0.1 MB/s, 1 MB/s, and 10 MB/s which are typical bandwidths in modern mobile communication networks.

Fig. \ref{fig:diff_bandwidth} illustrates the communication time of the proposed FedMS and other baselines on different datasets per communication round under different network bandwidths. Low bandwidth results in long data transmission time for parameter aggregation in each communication round thus impairs the training efficiency of FL.

From the results, we can observe that when the bandwidth is 0.1 MB/s, the time for FedMS to finish a communication round is 3.08 seconds while the communication time for FT, LFFT, and PromptFL per communication round is 5,984 seconds, 2,992 seconds, and 0.32 seconds. FedMS save $99.95\%$ communication time compared to the FT algorithm. When the bandwidth is 1 MB/s and 10 MB/s, the difference in training time between different algorithms is reduced. FedMS and PromptFL have communication times of fewer than 5 seconds per communication round in all network bandwidth conditions. Although PromptFL has the minimum communication resource consumption, both FedMS and PromptFL are communication efficient in real-world scenarios and FedMS has much higher accuracy on all datasets compared with PromptFL.

\subsubsection{Ablation study}
\begin{table}[]
\centering
\begin{tabular}{|c|c|}
\hline
Model type                & Number of parameters \\ \hline
Model with \texttt{MoFM} (Vit-B/16) & 311M            \\ \hline
Model without \texttt{MoFM} (Vit-L/14) & 428M            \\ \hline
\end{tabular}
    \caption{Comparison of the number of parameters of the model}
    \label{tab:parameter_ablation}
\end{table}

\textbf{Impact of Mixture of Foundation Models architecture.}
We compare the image classification accuracy of the model with the \texttt{MoFM} architecture using the visual encoder ViT-B/16 and the model without the \texttt{MoFM} architecture but with a visual encoder ViT-L/14, which has a much larger number of parameters. The first model is trained through all two stages of FedMS and the second model is trained through the first stage of FedMS. The total number of epochs of both models is set to 50 to ensure fair comparison. The comparison of the total number of parameters is shown in Tab. \ref{tab:parameter_ablation}. The model with \texttt{MoFM} architecture but with a smaller visual encoder has $27.33\%$ parameter less than another model.

We can observe from Fig. \ref{fig:Comparison_diff_vit} that although with much fewer parameters, the model with the \texttt{MoFM} architecture and trained through complete two stages achieves a higher accuracy on all datasets. It has an average accuracy increase of $2.36\%$ on all datasets. More specifically, with an increase of $2.03\%, 3.24\%$, and $1.82\%$ on Food101, UCF101, and EuroSAT datasets.

This surprising result demonstrates the superiority of the \texttt{MoFM} architecture as it can intelligently assign different weights to different experts according to the characteristics of the images to be classified. This enables better generalization ability to data that is out of the distribution of clients' local datasets and enables better personalization ability to data that follow the local distribution of the local clients' local datasets.

\textbf{Impact of aggregation of gate parameters.}
We compare the image classification accuracy of FedMS and FedMS with all gate parameters activated in the second training stage. We can observe from Tab. \ref{tab:gate_parameter_ablation} that by only fine-tuning the last layer of the gate adapter, we can achieve an accuracy of $93.73\%, 87.33\%$, and $99.46\%$ in Food101, UCF101, and EuroSAT datasets which is $0.09\%, 0.66\%$, and $0.35\%$ higher than tuning full parameters of the gate model. Moreover, we save the communication resource consumption by $97.69\%$. 

This is because the parameters changing of the local expert during training can cause the relationship between the global expert and the local expert to keep changing which makes it hard for the gate model to decide the decision weight it assigns to the two FMs and can cause unstableness in training which harm the performance. By freezing the gate parameters and inserting a lightweight gate adapter, the gate model can quickly adapt to the newly optimized parameters of the local expert while maintaining the feature extraction ability.


\begin{table}[]
  \centering

\begin{tabular}{|l|ll|}
\hline
        & \multicolumn{2}{c|}{Accuracy}                              \\ \cline{2-3} 
Datasets        & \multicolumn{1}{c|}{Gate adapter only} & All gate parameters \\ \hline
Food101 & \multicolumn{1}{c|}{93.73\%}           & 93.64\%           \\ \hline
UCF101  & \multicolumn{1}{c|}{87.33\%}           & 86.67\%           \\ \hline
EuroSAT & \multicolumn{1}{c|}{99.46\%}           & 99.11\%           \\ \hline
\end{tabular}
  \caption{Comparison of accuracy on different datasets}
\end{table}

        

\begin{table}[]
    \centering
\begin{tabular}{|c|c|}
\hline
Model types  & Number of parameters  \\ \hline
Gate adapter & 264.19k             \\ \hline
Gate model   & 11.44M             \\ \hline
\end{tabular}
    \caption{Comparison of the number of parameters of the gate adapter and the gate model}
    \label{tab:gate_parameter_ablation}
\end{table}



\section{Conclusions}

In this paper, we propose a novel FedMS algorithm that contains two training stages
to address the computation, communication, and statistical heterogeneous challenges in federated learning with foundation models. In the first stage, we freeze the pre-trained FM weight and insert low-rank decomposition matrices in every transformer block. We activate all the inserted matrices to better extract global feature information. In every communication round only the parameters of low-rank decomposition matrices join the weight aggregation. In the second stage, we take FM trained in the first stage as the global expert and construct another local expert to provide personalization for individual clients. We are the first to form the global expert and the local expert as a Mixture of Foundation Models (\texttt{MoFM}) in federated learning. We specially design and insert a gate adapter into the gate model to help assign the decision weight of the two experts. Moreover, to enable efficient training in computation-scarce scenarios, we propose a Sparsely
Activated LoRA (\texttt{SAL}) algorithm to activate the low-rank adaptation matrices progressively according to past accuracies in the Capability Queue. 

We test the performance of FedMS through extensive experiments in various settings, and results show that FedMS outperforms other SOTA baselines.

\ifCLASSOPTIONcaptionsoff
  \newpage
\fi



%


\bibliography{reference.bib}

%








\bibliographystyle{IEEEtran}

\end{document}